\documentclass{article} 
\usepackage{iclr2025_conference,times}
\usepackage{graphicx} 

\usepackage{amsmath,amsfonts,bm}

\def\eqref#1{equation~\ref{#1}}

\def\1{\bm{1}}

\DeclareMathAlphabet{\mathsfit}{\encodingdefault}{\sfdefault}{m}{sl}
\SetMathAlphabet{\mathsfit}{bold}{\encodingdefault}{\sfdefault}{bx}{n}

\usepackage{hyperref}
\usepackage{url}
\usepackage{booktabs}
\usepackage{makecell}
\usepackage{array}
\usepackage{adjustbox}
\usepackage{multirow}
\usepackage{siunitx}
\usepackage{enumitem}
\usepackage[utf8]{inputenc}
\usepackage{algorithm}
\usepackage{algpseudocode}
\usepackage{amsmath}
\usepackage{booktabs}
\usepackage[most]{tcolorbox}
\usepackage{amsmath}
\usepackage{lipsum}
\usepackage{enumitem}
\usepackage{subcaption}
\usepackage{cleveref}

\usepackage{listings}
\usepackage{xcolor}

\definecolor{light-yellow}{RGB}{255,255,224}
\definecolor{light-blue}{RGB}{230,230,250}
\definecolor{light-green}{RGB}{240,255,240}

\lstdefinestyle{yellowstyle}{
  backgroundcolor=\color{light-yellow},
  basicstyle=\footnotesize\ttfamily,
  breaklines=true,
  frame=single,
  framesep=5pt,
  framexleftmargin=5pt,
  framexrightmargin=5pt,
}

\lstdefinestyle{bluestyle}{
  backgroundcolor=\color{light-blue},
  basicstyle=\footnotesize\ttfamily,
  breaklines=true,
  frame=single,
  framesep=5pt,
  framexleftmargin=5pt,
  framexrightmargin=5pt,
}

\lstdefinestyle{greenstyle}{
  backgroundcolor=\color{light-green},
  basicstyle=\footnotesize\ttfamily,
  breaklines=true,
  frame=single,
  framesep=5pt,
  framexleftmargin=5pt,
  framexrightmargin=5pt,
}

\title{Scaling Instruction-Tuned LLMs to Million-Token Contexts via Hierarchical Synthetic Data Generation}

\author{%
  Linda He\textsuperscript{1}\thanks{Work done during an internship at Together AI.} \ \ 
  Jue Wang\textsuperscript{2} \ \ 
  Maurice Weber\textsuperscript{2} \ \ 
  Shang Zhu\textsuperscript{2} \ \ 
  Ben Athiwaratkun\textsuperscript{2} \ \ 
  Ce Zhang\textsuperscript{2,3}
  \\
  \textsuperscript{1}Harvard University \ \ 
  \textsuperscript{2}Together AI \ \ 
  \textsuperscript{3}University of Chicago \ \ 
  \\
  {\tt\small lindahe@college.harvard.edu, \{jue, maurice, shang, ben\}@together.ai,}
  \\
  {\tt\small cez@uchicago.edu}
}

\iclrfinalcopy
 
\begin{document}

\maketitle

\begin{abstract}
Large Language Models (LLMs) struggle with long-context reasoning, not only due to the quadratic scaling of computational complexity with sequence length but also because of the scarcity and expense of annotating long-context data.
There has been barely any open-source work that systematically ablates long-context data, nor is there any openly available instruction tuning dataset with contexts surpassing 100K tokens.
To bridge this gap, we introduce a novel post-training synthetic data generation strategy designed to efficiently extend the context window of LLMs while preserving their general task performance. 
Our approach scalably extends to arbitrarily long context lengths, unconstrained by the length of available real-world data, which effectively addresses the scarcity of raw long-context data. 
Through a step-by-step rotary position embedding (RoPE) scaling training strategy, 
we demonstrate that our model, with a context length of up to 1M tokens, performs well on the RULER benchmark and InfiniteBench and maintains robust performance on general language tasks.

\end{abstract}

\section{Introduction}

The capabilities of Large Language Models (LLMs) have significantly advanced, enabling impressive performance across a wide range of natural language processing tasks \citep{wu2023bloomberggptlargelanguagemodel,jiang2023mistral7b,wei2022emergentabilitieslargelanguage}. 
However, managing long contexts remains a major challenge, which limits the practical utility of LLMs in tasks such as document comprehension and summarization, code generation, lifelong conversations, and complex agent scenarios \citep{liu2023lostmiddlelanguagemodels,meng2023locatingeditingfactualassociations}.
Extending context lengths to 1M tokens marks a critical breakthrough for applications requiring processing beyond a 128K token limit. For instance, company-wide document retrieval benefits from efficiently analyzing extensive organizational histories stored in unstructured formats, while interconnected project timelines and legal documents gain from enhanced reasoning across multi-document datasets.

To extend the context length of LLMs, current approaches focus on either architectural innovations like efficient attention mechanisms ~\citep{katharopoulos2020transformersrnnsfastautoregressive,gu2024mambalineartimesequencemodeling} or scaling positional embeddings \citep{chen2023extendingcontextwindowlarge, peng2023yarnefficientcontextwindow} and continual pretraining on natural long-form data, such as books and web data. However, the RULER benchmark \citep{hsieh2024rulerwhatsrealcontext} shows that many models struggle to maintain consistent performance as context length increases, even when claiming to support longer contexts. 
This highlights the need for high-quality instruction data to fully utilize the nuances of long-form content. Acquiring such data is challenging and costly, as open-source datasets often fall short in document length, relevance, and tasks requiring genuine long-range understanding. 
To date, no open-source instruction-tuning datasets exceed 100K tokens, creating a significant gap between theoretical and practical long-context capabilities of LLMs \citep{li2024looglelongcontextlanguagemodels, zhao2024longskyworktrainingrecipeefficiently}.

\begin{figure}[t]
    \centering
    \includegraphics[width=.8\linewidth]{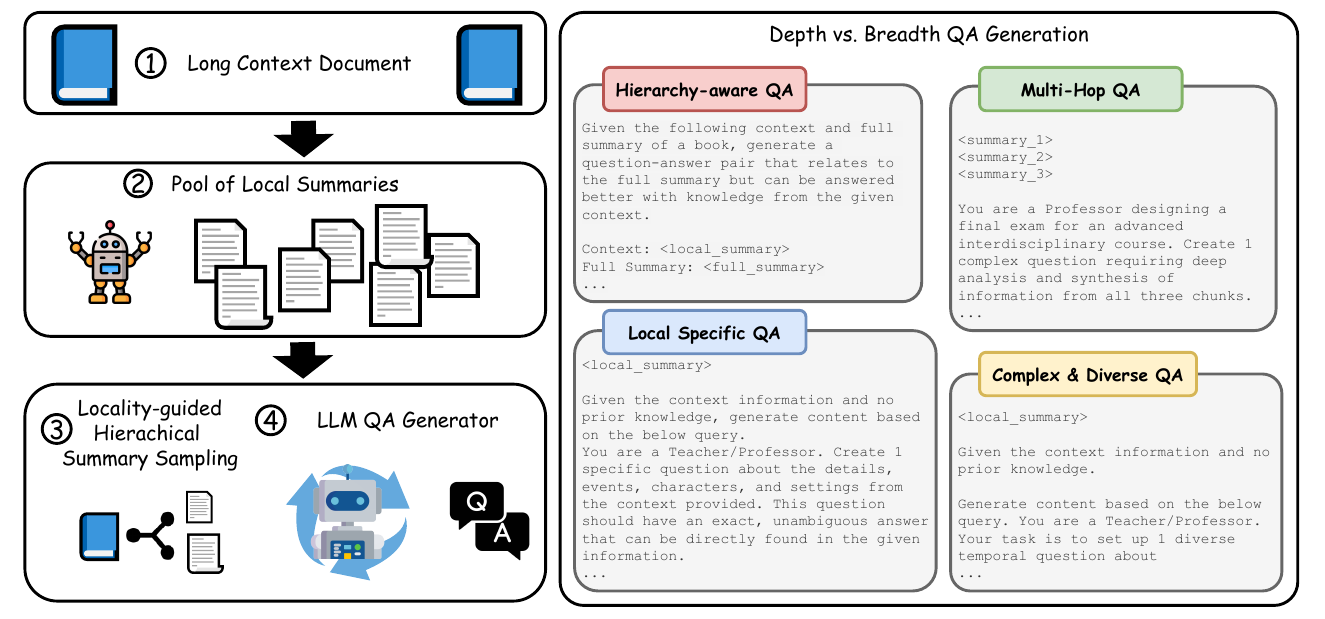}
    \caption{High-level overview over our approach to automatically generate QA pairs for long context documents. (1) In the first step, we split a document into small and medium chunks which are then (2) summarized by an off-the-shelf LLM requiring only smaller context windows. In (3) we sample summaries at different localities in a hierarchical manner, balancing local and global views of the original document. In (4) we generate questions based on the sampled summaries. In the right panel, we show a subset of prompts used to generate diverse and complex questions, given the sampled summaries.}
    \label{fig:single-doc}
    \vspace{-0.5cm}
\end{figure}

To address limitations in extending LLM context length, we propose an effective long-context instruction data generation pipeline, as illustrated in \Cref{fig:single-doc}. 
Our pipeline leverages short-context models to create long-context instruction data using three key methods: 
(a) \textit{Hierarchical question ordering}: 
structuring questions in a logical sequence to ensure coherent reasoning across contexts;
(b) \textit{Diverse question type pool}: 
maintaining a wide range of question types, including hierarchical-aware, multi-hop, local-specific, and other complex types to handle varied tasks; and (c) \textit{Multi-document integration}: incorporating multiple documents to generate data with arbitrary context lengths.
The contributions of this paper are threefold:

\begin{enumerate}[noitemsep, topsep=0pt, partopsep=0pt, parsep=0pt, left=0em]
    \item \underline{\textbf{Extensive and scalable long-context data generation strategy}}: 
    We present, to the best of our knowledge, the first extensive strategy for synthetically generating long-context data with comprehensive ablation tests and evaluations. Our highly scalable approach is unconstrained by the length of available real-world data, effectively combining multiple documents with diverse, complex questions. This hierarchical method ensures logical coherence and sequence integrity.
    
    \item \underline{\textbf{Extensive evaluation of core strategies}}: 
    We conduct extensive evaluations on shorter context lengths (100K and 180K) to demonstrate the effectiveness of our hierarchical strategy, multi-document combinations, and diverse question-answer pair generation. These evaluations validate that our core strategies work well across various tasks and context lengths.
    
    \item \underline{\textbf{Scaling to 1M context length}}: 
    We successfully extend LLaMA-3.1-8B-Instruct to a context length of 1 million tokens. Our model significantly outperforms the LLaMA-3.1-8B-Instruct model in zero-shot RoPE scaling to a 1M context window on the RULER benchmark and surpasses the gradientai/Llama-3-8B-Instruct-Gradient-1048k model trained by Gradient AI. Additionally, our model outcompetes LLaMA-3.1-8B-Instruct on InfiniteBench while maintaining strong performance on LongBench and MMLU.
\end{enumerate}

The remainder of this work is organized as follows. In \Cref{sec:related_work} we place our work in the landscape of existing literature around methods to address long context capabilities of LLMs. \Cref{sec:method} presents our method for generating long-context instruction tuning data. Our approach is then validated in \Cref{sec:experiments} with a series of extensive and representative experiments. Finally, we conclude in \Cref{sec:conclusion}. 

\section{Related Work}
\label{sec:related_work}
\vspace{-2mm}
Adapting transformers to enable longer context capabilities is a critical area of research in natural language processing. This effort primarily focuses on three key directions: (1) architectural modifications to the transformer model itself, (2) improvements in positional embedding techniques, and (3) the development and utilization of more extensive long-context datasets.

\textbf{Efficient Attention Mechanisms. }
To address the quadratic computational and memory demands of standard transformer self-attention, researchers have developed various architectural modifications to improve efficiency and extend context lengths. Notable examples include Longformer \citep{beltagy2020longformerlongdocumenttransformer},  which combines sliding window and global attention, and BlockTransformer \citep{ho2024blocktransformerglobaltolocallanguage}, which employs hierarchical global-to-local modeling. Linear Attention methods \citep{katharopoulos2020transformersrnnsfastautoregressive} reformulate self-attention for linear complexity, while InfiniteTransformer \citep{munkhdalai2024leavecontextbehindefficient} incorporates unbounded long-term memory through continuous-space attention. 
State space models like Mamba \citep{gu2024mambalineartimesequencemodeling} capture long-range dependencies efficiently without explicit attention mechanisms. Despite these advancements, bridging the gap with high-quality data remains a critical challenge and is the focus of this work.

\textbf{Position Embedding Extension.} 
Advances in positional encoding methods have enabled language models to handle longer sequences effectively. Techniques like RoPE \citep{su2023roformerenhancedtransformerrotary}, ALiBi \citep{press2022trainshorttestlong}, and xPos \citep{sun2022lengthextrapolatabletransformer} have emerged as prominent solutions. RoPE has gained widespread adoption in LLaMA \citep{touvron2023llamaopenefficientfoundation}, b) and PaLM \citep{anil2023palm2technicalreport}, due to its ability to represent relative positions and its theoretical grounding in the complex plane. A breakthrough showed that RoPE’s embeddings could extend to longer contexts with minimal or no fine-tuning  \citep{men2024baseropeboundscontext}, leading to two key approaches: Positional Interpolation (PI) \citep{chen2023extendingcontextwindowlarge} which linearly scales positional indices to extend context length, and NTK-aware Scaling RoPE \citep{peng2023yarnefficientcontextwindow} which combines high-frequency extrapolation with low-frequency interpolation. While these developments improve model performance with longer inputs, they rely heavily on limited long-context data for fine-tuning.

\textbf{Long Context Data.} 
Recent work, such as LongT5 \citep{guo2022longt5efficienttexttotexttransformer} and LongAlpaca \citep{chen2024longloraefficientfinetuninglongcontext}, has shown the benefits of additional pretraining on long sequences, enabling models to better capture extended context. Methods like combining multiple short-context sequences \citep{xiong2023effectivelongcontextscalingfoundation} have also emerged as promising ways to efficiently extend context lengths. However, a significant gap remains in generating high-quality instruction-tuning data exceeding 100K context lengths. Few open-source efforts address this need. Our work introduces a scalable pipeline for generating long-context instruction-tuning data by systematically combining multiple documents, diverse questions, and a hierarchical strategy to ensure coherence and structure. 

\textbf{Synthetic Data Generation.}
Synthetic data generation offers a promising path for scaling language models across diverse tasks and complex instructions. AutoEvol-Instruct \citep{zeng2024automaticinstructionevolvinglarge}, automates the evolution of instruction datasets using large language models, reducing the need for extensive human intervention. WizardLM  \citep{xu2023wizardlmempoweringlargelanguage} employs Evol-Instruct to iteratively evolve and scale instruction complexity, achieving strong results on benchmarks like MT-Bench and Vicuna’s evaluation set. Auto Evol-Instruct ~\citep{zeng2024automaticinstructionevolvinglarge} further refines this process with an iterative evolution strategy, while Self-Instruct \citep{wang2023selfinstructaligninglanguagemodels} enhances instruction-following performance through data synthesis. Our work extends this research by generating long-context data tailored for instruction tuning.

 \vspace{-2mm}
\section{Method}
\label{sec:method}
\vspace{-2mm}
In this section, we describe our methodology for generating coherent instructions from a single document and scaling it to multiple documents to curate long-context datasets beyond the context length of available raw data. Section \ref{sec:coherent-single} outlines our strategy for ensuring (1) \textit{quality and complexity} and (2) \textit{coherent ordering} of generated question-answer pairs. Section \ref{sec:multi-doc} expands on scaling to longer context lengths using multiple documents. \Cref{fig:single-doc} provides an overview of our long-context synthetic data generation pipeline.

\vspace{-3mm}
\subsection{Coherent Instructions from a Single Document}
\label{sec:coherent-single}
\vspace{-3mm}
The quality of long-context instruction-tuning datasets is driven by two key factors: (1) the \textit{complexity and diversity} of the generated instructions, and (2) the \textit{structured ordering} of questions and instructions. To address these, we devised a bifurcated strategy targeting each component. 


\textbf{Quality, Diversity, and Complexity of Instructions.}
As illustrated in Figure~\ref{fig:single-doc}, our methodology for generating rich, diverse, and complex instructions leverages the key insight that \textit{short-context models can be used to generate long-context instruction-tuning data}. The core approach involves dividing the input document into smaller chunks, typically 4K tokens, enabling models optimized for shorter contexts to process these segments with \textit{greater precision and clarity}. 
We curate an initial set of prompts covering multiple dimensions of instruction complexity, such as temporal reasoning, thematic inquiry, and character-based scenarios (full set in \Cref{apx:prompts}).
During question-answer pair generation, a small chunk and one question are randomly selected to generate a pair. To ensure broader contextual understanding, we incorporate multi-hop questions spanning 2–4 chunks, enabling cross-chunk question-answer pairs.

\begin{figure}[t]
    \vspace{-15mm}
    \centering
  \includegraphics[width=0.8\linewidth]{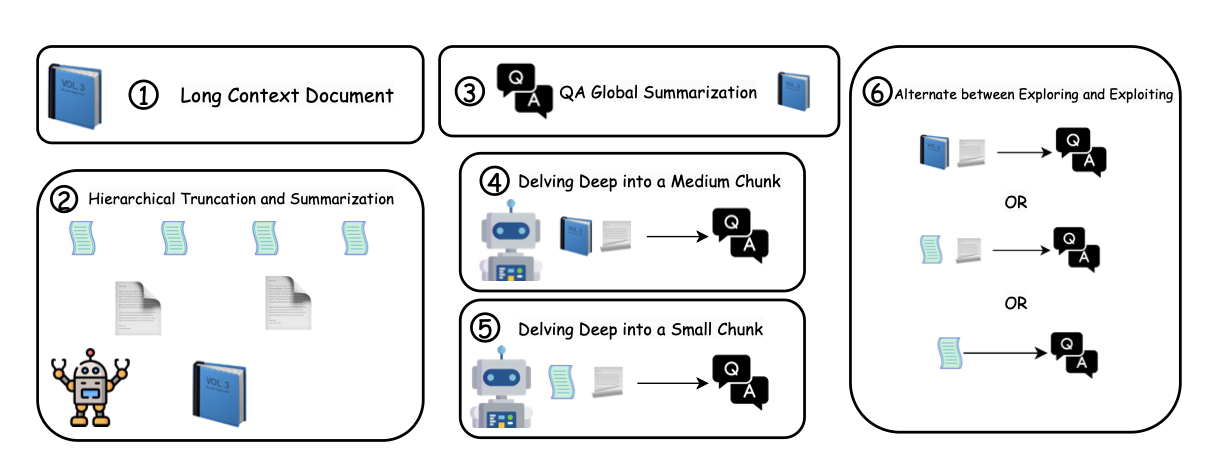}
    \caption{High-level overview over our approach to generate order-following QAs. (1) Input a raw long context document. (2) Split the document into small, medium, and global chunks, and generate summaries at each level. (3) The first QA is based on the global summary. (4) We randomly select a medium chunk to generate a QA, (5) then delve deeper by selecting a small chunk within it for another QA. (6) To continue, the process alternates between exploiting the same small chunk or exploring new medium or small chunks to generate further QAs.
}
    \label{fig:single-doc-order-following}
    \vspace{-2mm}
\end{figure}

\textbf{Ensuring Coherent Order.} To ensure logical and coherent QA generation, we use a hierarchical strategy to split, summarize, and generate questions from long documents (see Figure \ref{fig:single-doc-order-following}), balancing exploration and exploitation. 
The document is first divided into large sections of 12K tokens, then into smaller 4K-token chunks linked hierarchically to connect broader and granular segments.
The first QA is based on the global summary to give a high-level overview of the document. Then, we randomly select a medium chunk to generate a QA, and then delve deeper by selecting a small chunk within it for another QA. To continue, the process alternates between exploiting the same small chunk or exploring new medium or small chunks to generate further QAs. 
 This iterative process ensures a balance between specificity and diversity, covering both localized details and broader document sections. The hierarchical structure ensures logical progression from broad QAs to detailed ones. The detailed algorithm and pseudocode are provided in \Cref{sec:additional-details-algorithms}.

\begin{figure}[t]
    \centering
    \includegraphics[width=.9\linewidth]{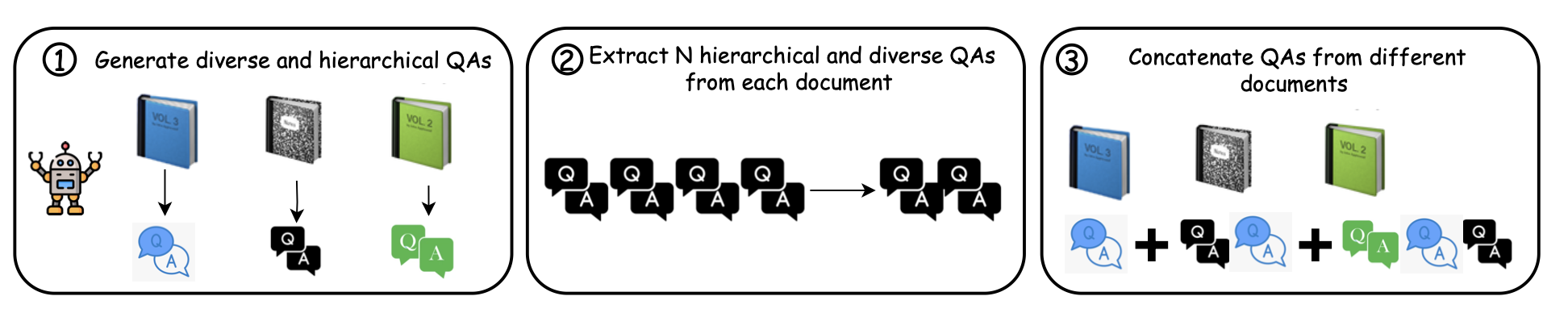}
    \caption{High-level overview over our approach to curate long context data using multiple documents. (1) Diverse and hierarchical QAs are generated at different levels of granularity for each document. (2) $N$ hierarchical and diverse QAs are sampled and extracted from each document. (3) QAs from different documents are combined, maintaining a balance of hierarchical and diverse questions across the entire set. $N = 5$ in our algorithm, and when we revisit previous documents in step (3), we sample 3 hierarchial questions for each document with 60 $\%$ probability as well as 9 total diverse questions from all previous documents. }\label{fig:multi-doc}
\end{figure}

\vspace{-2mm}
\subsection{Extending to Longer Context Lengths Using Multiple Documents}
\label{sec:multi-doc}
\vspace{-3mm}
Here we extend our methodology to handle longer contexts by concatenating multiple documents and generating coherent hierarchical and diverse QA pairs across them. The workflow is visualized in ~\Cref{fig:multi-doc} and the detailed algorithm is provided in \Cref{sec:additional-details-algorithms}. Below, we clearly define the parameters $N_1$, $N_2$, and $N_3$, which govern the selection of hierarchical and diverse QA pairs, ensuring logical continuity and broad reasoning across documents.
For each document, the process proceeds as follows:
\vspace{-2mm}
\begin{enumerate}
    \item $N_1$ hierarchical QA pairs and $N_1$ diverse QA pairs: After processing each document, $N_1 = 5$ hierarchical follow-up questions are added. These questions are designed to capture contextually related information within the document, creating a logical order of reasoning and flow across sections. Moreover, another $N_1 = 5$ diverse QA pairs for this document is added as well, designed to capture specific details of the document.
    \vspace{-5mm}
    \item $N_2$ diverse QA pairs: Next, $N_2 = 9$ diverse QA pairs are added. These questions are sampled from all previously visited documents where diverse QA pairs have not already been sampled. This approach ensures cross-referencing between documents.
    \vspace{-1mm}
    \item $N_3$ revisiting hierarchical QA pairs: For every previously visited document, there is a 60$\%$ probability of sampling $N_3 = 3$ hierarchical follow-up questions. These are added to revisit earlier contexts, fostering a richer and interconnected understanding of the content.
 
\vspace{-2mm}
\end{enumerate}

This process is repeated iteratively for all $K$ documents in the dataset to create a comprehensive instruction-tuning dataset that balances within-document reasoning, cross-document relationships, and revisiting earlier content for contextual continuity. We also present an example of a concatenated data example in \Cref{sec:additional-data-example}.



\vspace{-2mm}
\section{Experiments}
\label{sec:experiments}
\vspace{-2mm}
In this section, we validate our long-context data generation approach through a series of experiments. 
In Section \ref{sec:result-main}, we extend LLaMA-3.1-8B-Instruct to a 1M context-length model using stepwise RoPE scaling and hierarchical, diverse QA data generated by Qwen-2-72B. 
Our 1M model delivers excellent results on ultra-long contexts while maintaining strong performance on short and medium-length contexts.
In Section \ref{sec:llama-qwen-generators}, we evaluate robustness using smaller and same-sized generator models (Qwen-2.5-7B and LLaMA-3.1-8B-Instruct), confirming our models achieve strong performance across ultra-long, short, and medium contexts. These findings highlight the scalability and effectiveness of our approach across generator model sizes.
In Section \ref{sec:ablations}, we present ablation studies showing how our hierarchical strategy and diversified questions significantly improve long-context instruction tuning, focusing on 180K with two documents.




\vspace{-2mm}
\subsection{Setup}
\label{sec:setup}
\vspace{-2mm}


{\bf Models.} We use LLaMA-3.1-8B-Instruct as the base model for instruction-tuning, given its capability as a leading open-source LLM. To validate robustness, we employ various generator models for synthetic data: Qwen-2-72B-Instruct (large, high-quality data), Qwen-2.5-7B-Instruct (smaller), and LLaMA-3.1-8B-Instruct (same size). 
This demonstrates that our improvements are not reliant on very large models and that smaller models can achieve similar gains.
We also benchmark against the Gradient AI model (gradientai/Llama-3-8B-Instruct-Gradient-1048k), a 1M context-length model trained on 1.4 billion tokens, showing that our method outperforms existing baselines.

{\bf Hardware.} We fine-tuned our models on a SLURM cluster using 8 to 32 H100 GPUs across up to 4 nodes, connected via InfiniBand for efficient multinode training. We used FSDP to shard the model across GPUs and implemented DeepSpeed Ulysses sequence parallelism for long-context training.

{\bf Datasets.}  Our primary dataset is the Together long books dataset\footnote{\url{https://huggingface.co/datasets/togethercomputer/Long-Data-Collections}}, processed into approximately \textbf{1.4 billion} tokens, distributed across these stages: 2000 samples of 180K tokens, 1280 samples of 350K tokens, 600 samples of 650K tokens, and 200 samples of 1M tokens.
We generated 582,900 QA pairs with hierarchical and diverse questions for robust instruction-tuning using the Together AI inference API \footnote{\url{https://api.together.xyz/}}. 
By sending 32 simultaneous API requests, it took about two days to create our full long-context instruction dataset, comprising 7,772 books. For each book, we generated 25 hierarchical and 50 diverse questions, resulting in 582,900 QA pairs alongside global summaries.
During training, we calculate loss solely on answers, masking out questions and context to ensure the model focuses on reasoning and generating accurate answers without being penalized for reproducing input content.

{\bf Evaluation Protocol.} We evaluated our models using: 1) \textbf{InfiniteBench} \citep{zhang2024inftybenchextendinglongcontext}: Designed for LLMs on extended contexts, it includes tasks like key-value retrieval, summarization, and QA on data exceeding 100K tokens. We evaluated the first 150 samples per task, excluding coding tasks as our data lacks code. 2) \textbf{LongBench} \citep{bai2024longbenchbilingualmultitaskbenchmark}: Focused on medium-context tasks (10K tokens), it assesses summarization, QA, and fact-checking across multiple domains, offering a measure of general capabilities. We excluded coding tasks. 3) \textbf{RULER} \citep{hsieh2024rulerwhatsrealcontext}: RULER is a synthetic benchmark designed to evaluate how well models handle complex, real-world tasks in long contexts. Unlike traditional retrieval-based tasks like Needle-in-a-Haystack (NIAH), which focus on extracting specific pieces of information from distractor texts, RULER tests models' ability to comprehend deeper relationships and manage long-range dependencies. Given a specified context length, RULER generates synthetic tasks across multiple categories, including multi-hop reasoning and document tracing, and measures the model’s accuracy. In our evaluation, we sampled 130 tasks for each context length across 13 categories, totaling over 150 million tokens. 4) \textbf{MMLU} \citep{hendrycks2021measuringmassivemultitasklanguage}: This benchmark evaluates general model performance across multiple domains, assessing both breadth and depth of understanding. It includes tasks spanning STEM, humanities, and social sciences, with varying difficulty levels. MMLU ensures that improvements in handling long-context tasks do not cause regression in overall model capabilities.

\subsection{Main Results: Scaling Up to Longer Context Lengths (350K, 650K, 1M)}
\label{sec:result-main}
\vspace{-2mm}

\begin{figure}[t]
    \centering
    \begin{subfigure}[b]{0.32\linewidth}
        \centering
        \includegraphics[width=\linewidth]{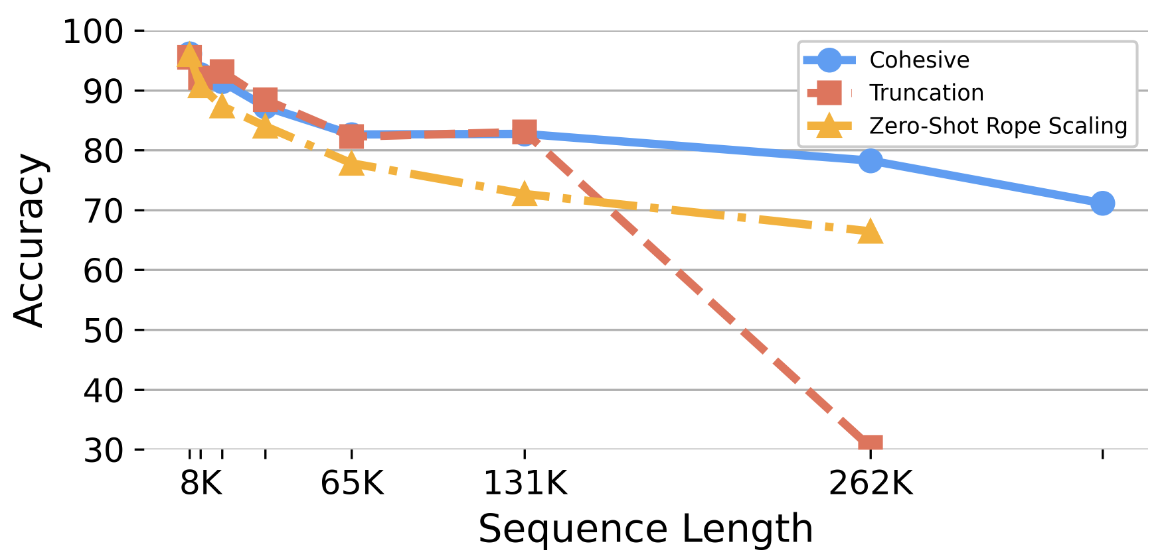}
        \caption{Context length 350K}
    \end{subfigure}
    \hfill
    \begin{subfigure}[b]{0.32\linewidth}
        \centering
        \includegraphics[width=\linewidth]{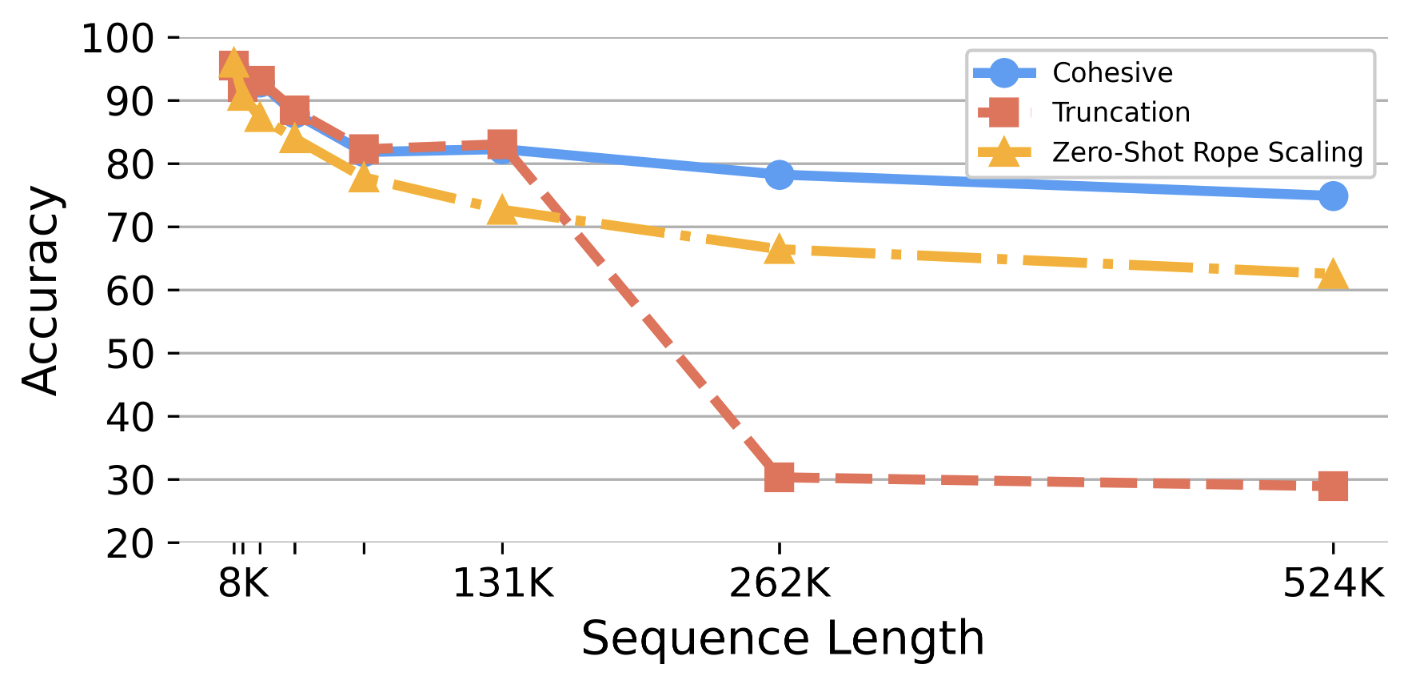}
        \caption{Context length 650K}
    \end{subfigure}
    \hfill
    \begin{subfigure}[b]{0.32\linewidth}
        \centering
        \includegraphics[width=\linewidth]{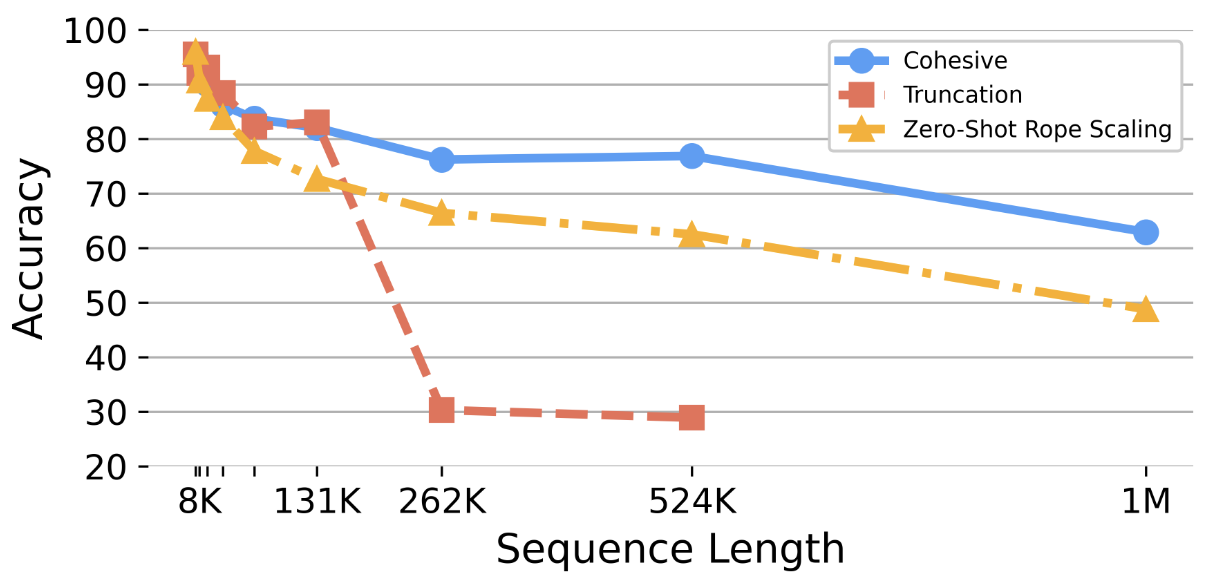}
        \caption{Context length 1M}
    \end{subfigure}
    \caption{
    \footnotesize Effective context length up 1M tokens using Qwen-2-72B-Instruct as generator on RULER.
    }
    \label{fig:main}
\end{figure}

To extend Llama-3.1-8B-Instruct to a 1M context model, we applied stepwise rope scaling. Training started with 180K tokens and progressed through checkpoints at 350K, 650K, and 1M tokens, concatenating 4, 8, and 12 documents as per the algorithm in Section \ref{sec:multi-doc}. 
We compiled 2000 samples at 180K, 1280 at 350K, 600 at 650K, and 200 at 1M tokens. Data was generated using Qwen-2-72B, fine-tuned on Llama-3.1-8B-Instruct with rope scaling at a 6e-5 learning rate for 1 epoch. Training the 650K model took 30 hours, and the 1M model took an additional 52.5 hours.

An earlier ablation test combining two documents (Section \ref{sec:ablations}) showed that combining hierarchical and diverse questions with a fixed number of QAs and global summarization is optimal for handling long contexts. We extended this setup for ultra-long context data, with each document followed by $N_1 = 5$ hierarchical and $N_1 = 5$ diverse questions. When revisiting previous documents, there is a 60$\%$ chance of extracting $N_2 = 3$ hierarchical question from each document and $N_3 = 9$ diverse questions sampled from all prior documents.

 \begin{table}[t]
\centering
\small

\caption{\footnotesize Model performance on InfiniteBench (100K tokens) using Qwen-2-72B-Instruct as generator.}
\label{tab:LLaMA-performance}

\begin{tabular}{@{}lcccccc@{}}
\toprule
\textbf{Metric} & \textbf{\makecell{LLaMA-3.1-\\8B-Instruct}} & \textbf{\makecell{gradient-\\ai-model}} & \textbf{180K} & \textbf{350K} & \textbf{650K} & \textbf{1M} \\
\midrule
Retrieve.PassKey & \textbf{100.00} & \textbf{100.00} & \textbf{100.00} & \textbf{100.00} & \textbf{100.00} & \textbf{100.00} \\
Retrieve.Number & 95.33 & 99.83 & 99.33 & \textbf{100.00} & \textbf{100.00} & \textbf{100.00} \\
Retrieve.KV     & 42.66 & 15.60 & 88.66 & \textbf{92.00} & 63.33 & 57.33 \\
En.Sum          & \textbf{27.63} & 17.02 & 24.01 & 23.51 & 23.68 & 23.06 \\
En.QA           & 24.83 & 14.31 & \textbf{34.26} & 33.23 & 31.72 & 31.97 \\
En.MC           & 68.00 & 57.20 & 74.00 & 72.00 & \textbf{75.33} & 74.00 \\
En.Dia          & 16.66 & 5.00 & 18.00 & 18.00 & \textbf{22.00} & 16.00 \\
Math.Find       & 35.33 & 19.42 & \textbf{37.33} & 35.33 & 36.00 & 36.00 \\
\midrule
\textbf{Average}& 51.31 & 41.04 & \textbf{59.45} & 59.26 & 56.51 & 54.80 \\
\bottomrule
\end{tabular}
\label{table:main-infini}
\end{table}


\begin{table}[t]
\centering
\small
\caption{Model performance on LongBench (10K tokens) using Qwen-2-72B-Instruct as generator.}
\label{tab:model-performance-longbench}

\begin{tabular}{@{}lcccccc@{}}
\toprule
 & {\textbf{\makecell{LLaMA-3.1-\\8B-Instruct}}} & {\textbf{\makecell{Gradient-\\AI-Model}}} & {\textbf{180K}} & {\textbf{350K}} & {\textbf{650K}} & {\textbf{1M}} \\
\midrule
Single Document & \textbf{46.91} & 30.71 & 45.83 & 45.88 & 45.24 & 45.15 \\
\midrule
Multi-Document & 41.45 & 12.45 & 41.71 & \textbf{41.75} & 41.13 & 41.29 \\
\midrule
Summarization & \textbf{26.10} & 21.72 & 25.14 & 24.97 & 24.26 & 24.98 \\
\midrule
Few-shot Learning & \textbf{63.48} & 59.69 & 62.22 & 61.66 & 60.00 & 59.27 \\
\midrule
Synthetic Tasks & 67.48 & 55.50 & \textbf{68.17} & 67.50 & 65.00 & 66.42 \\
\midrule
All & \textbf{48.11} & 35.89 & 47.58 & 47.34 & 46.18 & 46.42 \\
\bottomrule
\end{tabular}

\label{table:longbench}
\end{table}

\begin{table}[t]
\centering
\small

\caption{Model performance on MMLU using Qwen-2-72B-Instruct as the generator.}
\label{tab:model-comparison}
\begin{tabular}{@{}lccccc@{}}
\toprule
\textbf{Category}       & \textbf{\makecell{LLaMA-3.1-\\8B-Instruct}} & \textbf{\makecell{gradient-\\ai-model}} & \textbf{350K-model} & \textbf{650K-model} & \textbf{1M-model} \\
\midrule
mmlu                    & 68.21 $\pm$ 0.37              & 60.48 $\pm$ 0.39           & 66.29 $\pm$ 0.38    & 65.80 $\pm$ 0.38    & 65.08 $\pm$ 0.38  \\
humanities              & 64.23 $\pm$ 0.67              & 55.75 $\pm$ 0.69           & 61.51 $\pm$ 0.68    & 61.02 $\pm$ 0.68    & 61.02 $\pm$ 0.68  \\
other                   & 73.03 $\pm$ 0.77              & 67.04 $\pm$ 0.82           & 72.84 $\pm$ 0.77    & 71.84 $\pm$ 0.78    & 71.84 $\pm$ 0.78  \\
social sciences         & 77.48 $\pm$ 0.74              & 70.46 $\pm$ 0.80           & 76.81 $\pm$ 0.74    & 75.27 $\pm$ 0.76    & 75.27 $\pm$ 0.76  \\
stem                    & 60.36 $\pm$ 0.83              & 51.32 $\pm$ 0.86           & 59.44 $\pm$ 0.84    & 57.72 $\pm$ 0.84    & 57.72 $\pm$ 0.84  \\
\bottomrule
\end{tabular}
\label{table:mmlu}
\end{table}


Figure \ref{fig:main} shows the effective context lengths of the 350K, 650K, and 1M models on the RULER benchmark.
For comparison, we performed zero-shot rope scaling on the LLaMA-3.1-8B-Instruct model and included results using input truncation for context lengths above 128K as an additional baseline.
On contexts shorter than 128K, our models performed comparably to LLaMA-3.1-8B-Instruct and surpassed it with zero-shot rope scaling. This demonstrates the robustness of our models on short and medium contexts. For contexts longer than 128K, our models significantly outperformed both baselines, with their strengths becoming more evident as context length increased. Raw evaluation results are in Appendix \ref{sec:ruler-numerical}.

To further validate our approach, we compared it to the Gradient AI model (gradientai/Llama-3-8B-Instruct-Gradient-1048k), a 1M context model, on InfiniteBench, LongBench, and MMLU benchmarks.
Table \ref{table:main-infini} compares models across context lengths on InfiniteBench, while Table \ref{table:longbench} focuses on LongBench.
All our models (180K, 350K, 650K, 1M) consistently outperforms the Gradient AI model on InfiniteBench, showcasing the effectiveness of our hierarchical, diversified QA-based data-generation strategy.
The 180K and 350K models scored 59.45 and 59.26, significantly exceeding the LLaMA-3.1-8B-Instruct baseline of 51.31. The 650K model scored 56.51, and the 1M model achieved a strong 54.80. \footnote{The results dropped likely due to multi-node training, as we believe our 650K and 1M models are under-trained because of the extended time required to train and the communication overhead from NCCL.}

Notably, while the Retrieve.KV task shows the most significant improvement, tasks like Retrieve.Number, En.MC, and Math.Find also display meaningful gains. The improvement on Retrieve.KV stems from our data-generation methodology, which uses a structured mix of hierarchical and diverse questions while revisiting prior documents. This encourages the model to associate relevant sections, aligning with the demands of key-value retrieval and RAG techniques, where accurate context memorization is critical. Beyond key-value retrieval, our model excels on other tasks: on En.MC, the 650K model scored 75.33, surpassing the baseline (68.00) and Gradient AI model (57.20). On Math.Find, it scored 36.00 at 650K, outperforming the Gradient AI model (19.42), showcasing improved reasoning capabilities.

As shown in Table \ref{table:longbench}, , our models maintain robust short-context performance on LongBench, despite being trained for significantly longer contexts (up to 1M tokens). For example, our 1M context-length model achieves an average score of 46.42, comparable to the baseline LLaMA-3.1-8B-Instruct model (48.11). This demonstrates that while optimized for ultra-long contexts, the model generalizes effectively to shorter contexts, such as those on LongBench.
Minor regressions in tasks like summarization are due to trade-offs when training for extended contexts. As the model adapts to handle extremely long contexts, small task-specific adjustments may impact short-context performance. However, these regressions are minimal and expected, given the differences between short- and long-context tasks. Despite these trade-offs, our model consistently outperforms the Gradient AI model (35.89) on all LongBench tasks, demonstrating the effectiveness of our hierarchical and diversified instruction-tuning approach.

As detailed in Table \ref{table:mmlu}, our model demonstrated minimal regression in general task performance despite significant improvements in ultra-long-context tasks. For instance, our model retained competitive MMLU scores (e.g., 68.21 ± 0.37 for the baseline and 65.08 ± 0.38 for the 1M model), whereas the Gradient AI model showed marked degradation on both MMLU and LongBench. This reinforces the robustness of our method, ensuring that gains in ultra-long-context performance do not compromise broader capabilities.
In conclusion, our models excel at ultra-long-context tasks on RULER and InfiniteBench, outperforming the base LLaMA-3.1-8B-Instruct and Gradient AI models while maintaining strong performance on general tasks like MMLU and LongBench.

\vspace{-3mm}
\subsection{Validating Robustness Across Generator Models}
\label{sec:llama-qwen-generators}
\vspace{-3mm}
To validate that observed improvements are not solely due to using a large generator model (e.g., Qwen-2-72B), we trained and evaluated models with Qwen-2.5-7B and LLaMA-3.1-8B-Instruct as generators. By employing smaller or similarly sized models, we demonstrated the robustness and generalizability of our hierarchical QA data-generation strategy. Additionally, we benchmarked against the Gradient AI model (gradientai/Llama-3-8B-Instruct-Gradient-1048k), a 1M context model trained on 1.4 billion tokens. While our models were trained only up to 650K tokens to validate the approach, the same method can seamlessly scale to 1M tokens. Our models outperformed the Gradient AI baseline across all long-context benchmarks, achieving higher accuracy on InfiniteBench and RULER, while preserving general task performance on MMLU and LongBench.

\begin{figure}[t]
    \centering
    \begin{subfigure}[b]{0.45\linewidth}
        \centering
        \includegraphics[width=\linewidth]{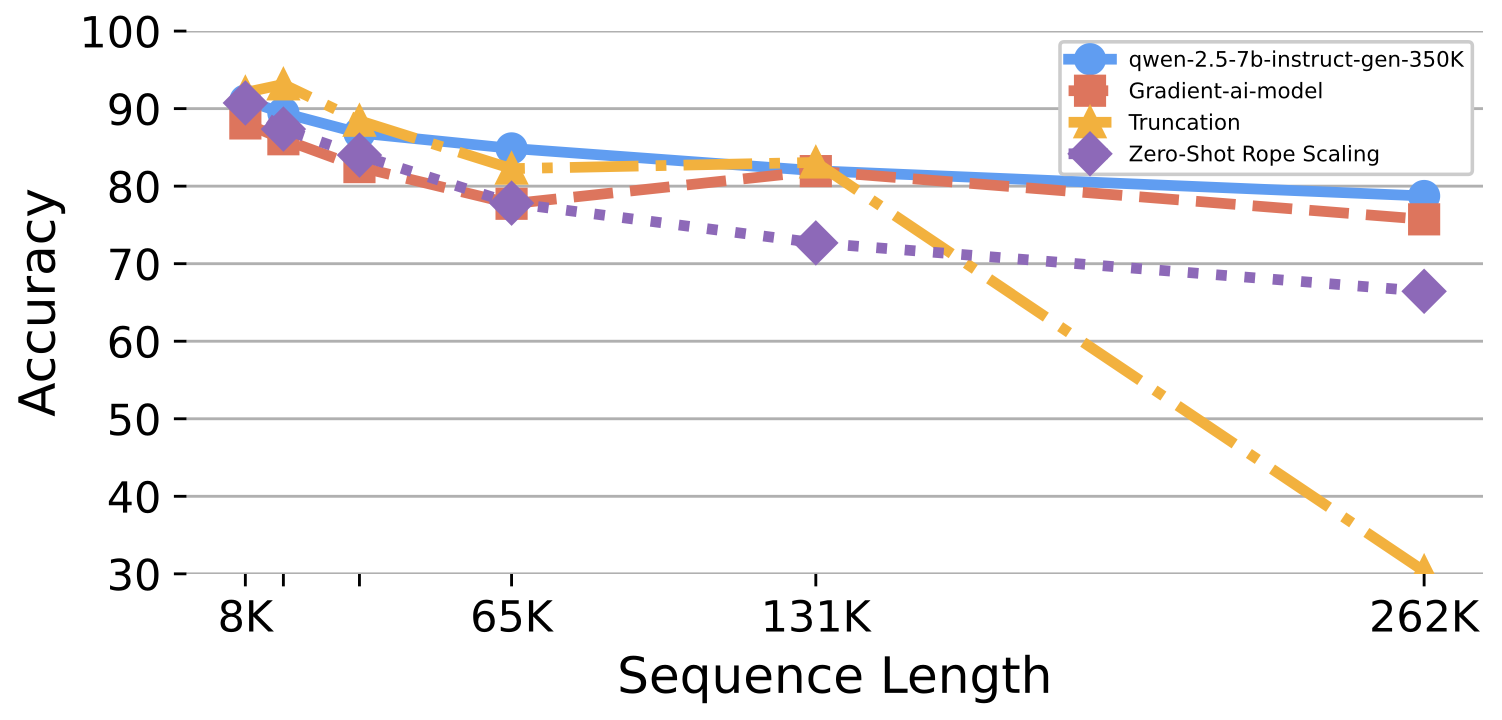}
        \caption{\scriptsize 350K model using Qwen-2.5-7B-Instruct as generator.}
    \end{subfigure}
    \hfill
    \begin{subfigure}[b]{0.45\linewidth}
        \centering
        \includegraphics[width=\linewidth]{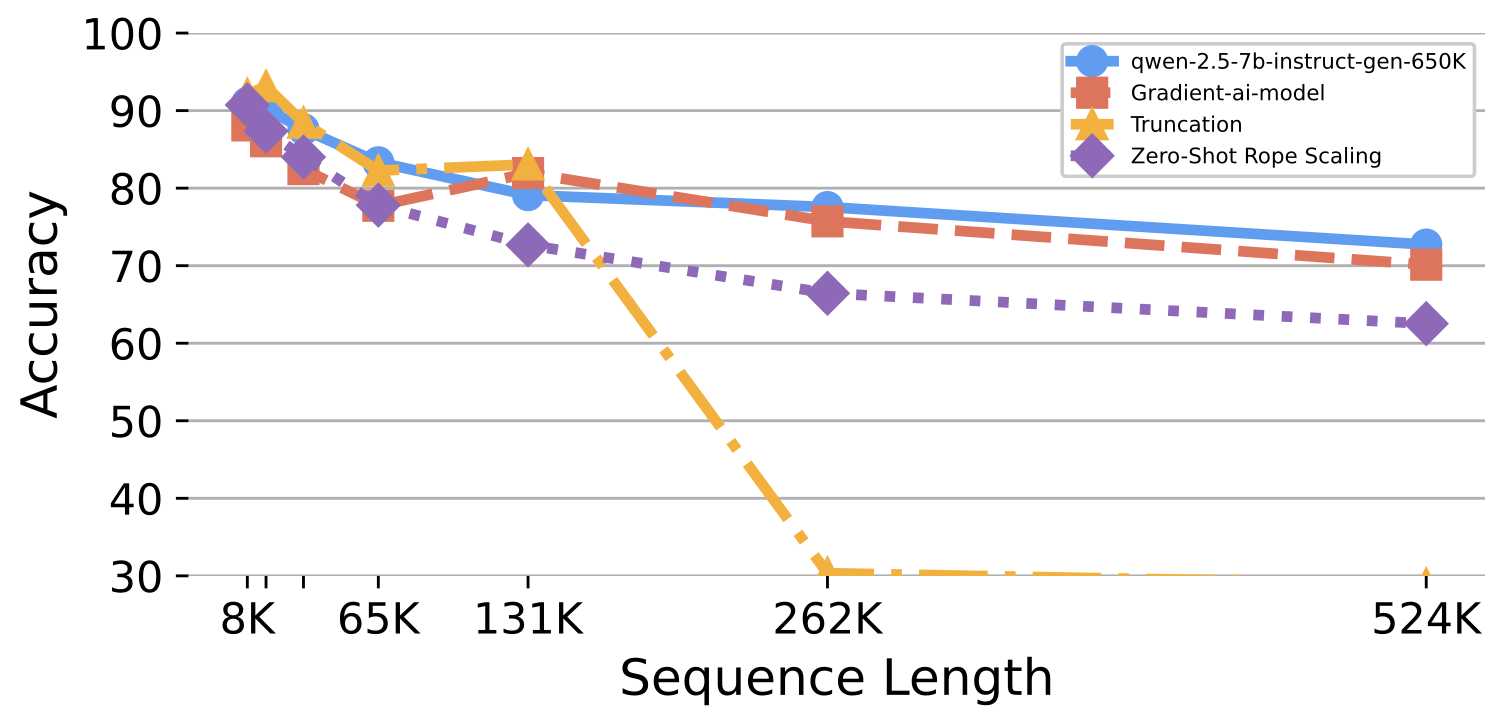}
        \caption{\scriptsize 650K model using Qwen-2.5-7B-Instruct as generator.}
    \end{subfigure}
    \hfill
    \begin{subfigure}[b]{0.45\linewidth}
        \centering
        \includegraphics[width=\linewidth]{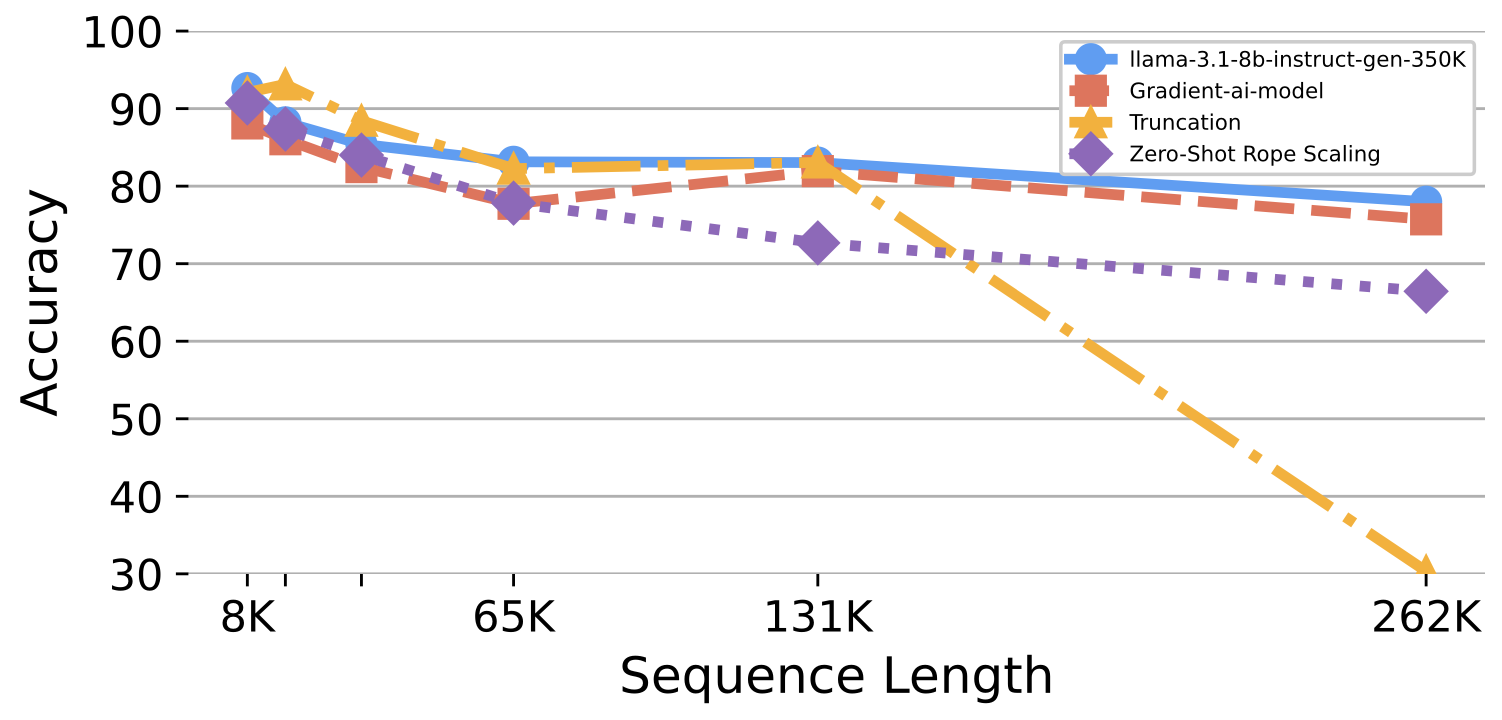}
        \caption{\scriptsize 350K model using Llama-3.1-8B-Instruct as generator.}
    \end{subfigure}
    \hfill
    \begin{subfigure}[b]{0.45\linewidth}
        \centering
        \includegraphics[width=\linewidth]{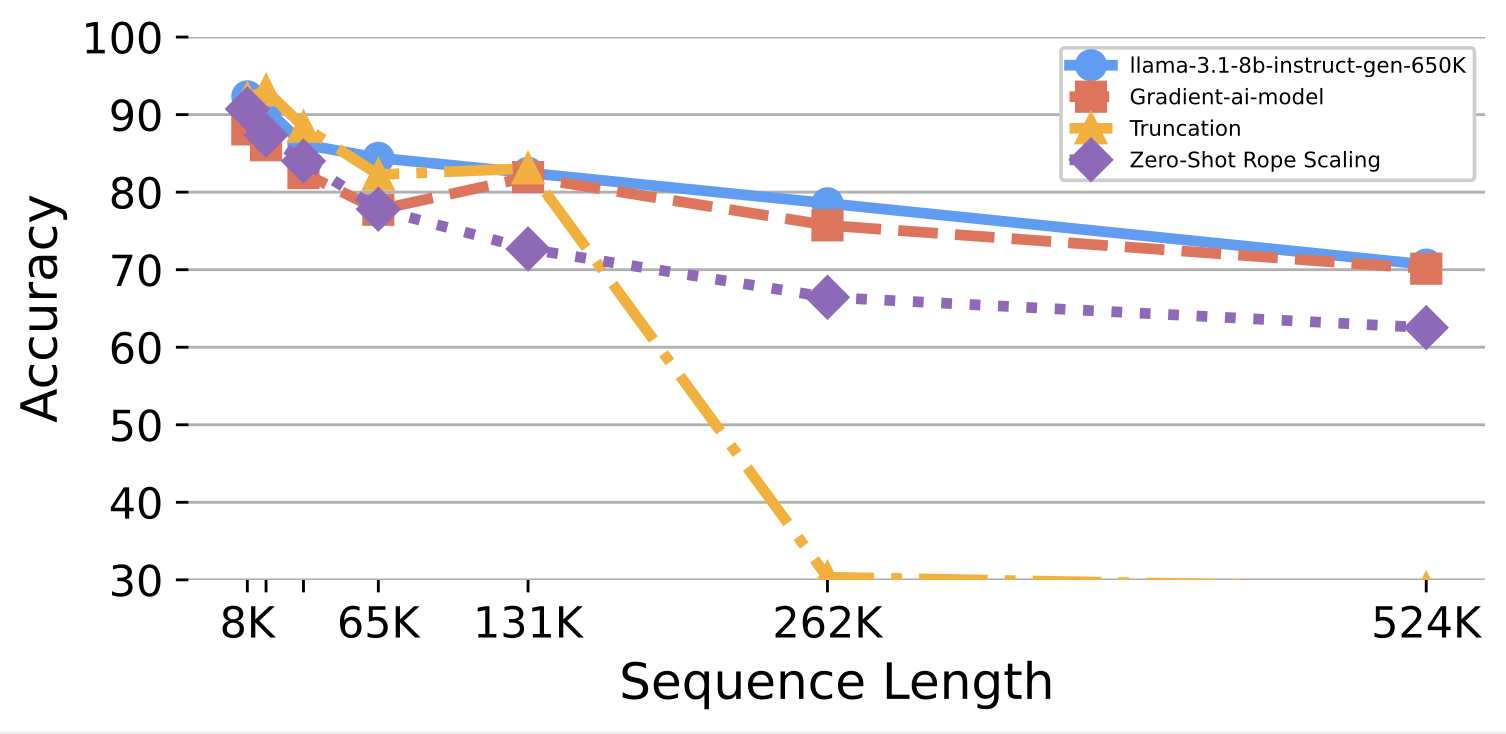}
        \caption{ \scriptsize 650K model using Llama-3.1-8B-Instruct as generator.}
    \end{subfigure}
    \caption{ \footnotesize
    Effective context length using Llama-3.1-8B-Instruct and Qwen-2.5-7B-Instruct as generators on RULER.
    }
    \label{fig:ruler-gen}
    \vspace{-4mm}
\end{figure}

Figure \ref{fig:ruler-gen} highlights effective context length using Llama-3.1-8B-Instruct and Qwen-2.5-7B as generators on RULER. On all settings (350K, 650K), our hierarchical approach outperformed the Gradient AI model and the zero-shot baselines across context lengths.
Table \ref{table:infinitebench-gen} summarizes results on InfiniteBench (100K context length). Our approach again consistently outperformed both the base LLaMA-3.1-8B-Instruct model and the Gradient AI model. 
This demonstrates that even smaller generator models produce high-quality data for instruction-tuning.

\begin{table}[t]
\centering
\scriptsize

\caption{ \footnotesize InfiniteBench performance with Llama-3.1-8B-Instruct and Qwen-2.5-7B-Instruct as generators.}
\label{tab:LLaMA-performance}
\begin{tabular}{@{}lcccccccc@{}}
\toprule
\textbf{Task} & \textbf{\makecell{LLaMA-3.1-\\8B-Instruct}} & \textbf{\makecell{gradient-\\ai-model}} & \textbf{\makecell{180K-\\llama-gen}} & \textbf{\makecell{350K-\\llama-gen}} & \textbf{\makecell{650K-\\llama-gen}} & \textbf{\makecell{180K-\\qwen-gen}} & \textbf{\makecell{350K-\\qwen-gen}} & \textbf{\makecell{650K-\\qwen-gen}} \\
\midrule
Retrieve.PassKey & 100.00 & 100.00 & 100.00 & 100.00 & 100.00 & 100.00 & 100.00 & 100.00 \\
Retrieve.Number & 95.33 & 99.33 & 99.04 & 100.00 & 100.00 & 99.76 & 100.00 & 100.00 \\
Retrieve.KV & 42.66 & 13.33 & 85.47 & 89.33 & 42.14 & 89.52 & 85.33 & 52.66 \\
En.Sum & 27.63 & 17.02 & 25.68 & 26.85 & 26.64 & 26.97 & 27.70 & 26.74 \\
En.QA & 24.83 & 15.84 & 33.39 & 35.67 & 33.37 & 32.30 & 29.55 & 29.67 \\
En.MC & 68.00 & 61.33 & 58.00 & 60.66 & 66.00 & 63.33 & 61.33 & 64.66 \\
En.Dia & 16.66 & 4.00 & 19.50 & 14.66 & 20.00 & 27.33 & 21.33 & 23.33 \\
Math.Find & 35.33 & 26.66 & 36.66 & 32.66 & 35.33 & 30.00 & 34.66 & 38.00 \\
\midrule
\textbf{Average} & 51.31 & 42.19 & 57.22 & 57.48 & 52.94 & 58.65 & 57.49 & 54.38 \\
\bottomrule
\end{tabular}
\label{table:infinitebench-gen}
\end{table}

\begin{table}[t]
\centering
\scriptsize

\caption{\footnotesize LongBench performance with Llama-3.1-8B-Instruct and Qwen-2.5-7B-Instruct as generators.}
\label{tab:model-performance}
\begin{tabular}{@{}lcccccccc@{}}
\toprule
\textbf{Task} & \textbf{\makecell{LLaMA-3.1-\\8B-Instruct}} & \textbf{\makecell{gradient-\\ai-model}} & \textbf{\makecell{180K-\\llama-gen}} & \textbf{\makecell{350K-\\llama-gen}} & \textbf{\makecell{650K-\\llama-gen}} & \textbf{\makecell{180K-\\qwen-gen}} & \textbf{\makecell{350K-\\qwen-gen}} & \textbf{\makecell{650K-\\qwen-gen}} \\
\midrule
single-document & 46.91 & 30.75 & 46.48 & 46.64 & 46.53 & 46.20 & 46.70 & 46.28 \\
multi-document & 41.45 & 12.45 & 38.69 & 38.75 & 37.54 & 40.76 & 41.90 & 39.31 \\
summarization & 26.10 & 21.72 & 25.28 & 25.10 & 24.68 & 25.05 & 24.83 & 24.90 \\
few-shot learning & 63.48 & 59.70 & 61.56 & 62.79 & 60.50 & 61.92 & 61.56 & 60.69 \\
synthetic tasks & 67.48 & 55.50 & 66.17 & 67.75 & 66.00 & 67.11 & 67.60 & 67.10 \\
\midrule
\textbf{Average} & 48.11 & 35.89 & 47.23 & 47.72 & 46.20 & 47.95 & 47.97 & 47.00 \\
\bottomrule
\end{tabular}
\label{table:longbench-gen}
\vspace{-0.2cm}
\end{table}

Table \ref{table:longbench-gen} evaluates model performance on LongBench (10K context length). Despite being optimized for ultra-long contexts, our approach retains strong performance on shorter contexts, comparable to LLaMA-3.1-8B-Instruct. For example, with Qwen-2.5-7B-Instruct as the generator, our model scored 47.00 at 650K, closely matching LLaMA-3.1-8B-Instruct's 48.11. Our model also outperforms Gradient AI (35.89) across all LongBench tasks. Table 6 shows our models' minimal regression in MMLU performance. The 650K trained using LLaMA-3.1-8B-Instruct as generator scored 65.93 ± 0.38, close to LLaMA-3.1-8B-Instruct (68.21 ± 0.37). In contrast, Gradient AI showed notable regression. This underscores our hierarchical approach’s ability to support long-context learning while maintaining general task performance.

\begin{table}[t]
\centering
\tiny

\caption{\footnotesize MMLU performance with Llama-3.1-8B-Instruct and Qwen-2.5-7B-Instruct as generators.}
\label{tab:model-comparison}
\begin{tabular}{@{}lcccccccc@{}}
\toprule
\textbf{Category} & \textbf{\makecell{LLaMA-3.1-\\8B-Instruct}} & \textbf{\makecell{gradient-\\ai-model}} & \textbf{\makecell{180K-\\llama-gen}} & \textbf{\makecell{350K-\\llama-gen}} & \textbf{\makecell{650K-\\llama-gen}} & \textbf{\makecell{180K-\\qwen-gen}} & \textbf{\makecell{350K-\\qwen-gen}} & \textbf{\makecell{650K-\\qwen-gen}} \\
\midrule
mmlu            & 68.21 $\pm$ 0.37 & 60.48 $\pm$ 0.39 & 66.99 $\pm$ 0.38 & 66.74 $\pm$ 0.38 & 65.93 $\pm$ 0.38 & 67.33 $\pm$ 0.38 & 65.78 $\pm$ 0.38 & 64.60 $\pm$ 0.38 \\
humanities      & 64.23 $\pm$ 0.67 & 55.75 $\pm$ 0.69 & 62.32 $\pm$ 0.67 & 61.38 $\pm$ 0.68 & 60.57 $\pm$ 0.68 & 62.81 $\pm$ 0.67 & 59.68 $\pm$ 0.68 & 59.45 $\pm$ 0.68 \\
other           & 73.03 $\pm$ 0.77 & 67.04 $\pm$ 0.82 & 72.90 $\pm$ 0.77 & 73.03 $\pm$ 0.76 & 72.87 $\pm$ 0.76 & 73.51 $\pm$ 0.76 & 73.00 $\pm$ 0.76 & 73.45 $\pm$ 0.77 \\
social sciences & 77.48 $\pm$ 0.74 & 70.46 $\pm$ 0.80 & 76.70 $\pm$ 0.74 & 76.93 $\pm$ 0.74 & 75.53 $\pm$ 0.75 & 76.76 $\pm$ 0.74 & 75.66 $\pm$ 0.75 & 71.87 $\pm$ 0.77 \\
stem            & 60.36 $\pm$ 0.83 & 51.32 $\pm$ 0.86 & 58.67 $\pm$ 0.84 & 58.61 $\pm$ 0.84 & 57.72 $\pm$ 0.84 & 58.77 $\pm$ 0.84 & 58.14 $\pm$ 0.84 & 56.49 $\pm$ 0.85 \\
\bottomrule
\end{tabular}
\label{table:mmlu-gen}
\end{table}

\subsection{Ablation Studies}
\label{sec:ablations}
\vspace{-2mm}
Our 100K context length single-document ablation studies, detailed in Appendix \ref{sec:ablation-100K}, demonstrate that hierarchical ordering significantly boosts performance, particularly when combined with diverse question sets. Configurations with hierarchical ordering consistently outperformed those without, highlighting its importance for structuring instruction-tuning data. These findings provide a solid foundation for extending our experiments to larger context lengths and exploring the interaction of hierarchical and diverse question compositions. Building on these results, we expanded our experimentation to a 180K context length combining two documents, aiming to determine whether the patterns observed at 100K scale effectively with rope scaling. We also explore which question types (hierarchical or diverse and complex) perform best for questions directly following documents or referencing previous ones.

\begin{table}[t]
\centering
\small
\caption{Ablation study on InfiniteBench with 180K context length. Each experiment is labeled as x-y-z, where x is the type of question after the first document, y is the type of question after the second document, and z is the type of question referencing after the second document is processed. For example, h-h-s-fixed is the dataset with 6 hierarchical questions following the first document, 6 hierarchical questions following the second document, and 4 follow-up diverse questions referencing the first document after the second document is processed. Randomized signifies that the number of questions sampled is randomized, and no-sum signifies that the global summary is removed.}
\label{tab:model-performance-two-rows}
\begin{tabular}{@{}lcccc@{}}
\toprule
\textbf{Task} & \textbf{\makecell{LLaMA-3.1-\\8B-Instruct}} & \textbf{\makecell{hs-hs-hs-\\randomized}} & \textbf{\makecell{hs-hs-hs-\\fixed}} & \textbf{\makecell{h-h-s-\\randomized}} \\
\midrule
Retrieve.PassKey & \textbf{100.00} & \textbf{100.00} & \textbf{100.00} & \textbf{100.00} \\
Retrieve.Number & 95.33 & \textbf{100.00} & 99.33 & \textbf{100.00} \\
Retrieve.KV     & 42.66 & 82.66  & \textbf{88.66} & 84.66 \\
En.Sum          & \textbf{27.63} & 23.42  & 24.01 & 24.33 \\
En.QA           & 24.83 & 33.32  & \textbf{34.26} & 31.84 \\
En.MC           & 68.00 & 71.33  & \textbf{74.00} & 73.33 \\
En.Dia          & 16.66 & 18.00  & 18.00 & 14.00 \\
Math.Find       & 35.33 & \textbf{39.33}  & 37.33 & 36.66 \\
\midrule
Average         & 51.31 & 58.51 & \textbf{59.45} & 58.10 \\
\midrule
\textbf{Task} & \textbf{\makecell{h-h-s-fixed-\\no-sum}} & \textbf{\makecell{h-h-s-\\fixed}} & \textbf{\makecell{h-h-\\randomized}} & \textbf{\makecell{h-h-h-\\randomized}} \\
\midrule
Retrieve.PassKey & \textbf{100.00} & \textbf{100.00} & \textbf{100.00} & \textbf{100.00} \\
Retrieve.Number  & 99.33 & 99.33 & 98.66 & 99.33 \\
Retrieve.KV      & 84.00 & 83.33 & 76.66 & 84.66 \\
En.Sum           & 24.11 & 24.74 & 24.33 & 23.86 \\
En.QA            & 32.81 & 33.88 & 30.69 & 31.97 \\
En.MC            & 70.66 & 73.33 & 72.00 & 72.00 \\
En.Dia           & 16.66 & 14.66 & 15.33 & \textbf{18.00} \\
Math.Find        & 36.66 & \textbf{39.33} & 35.33 & 35.33 \\
\midrule
Average         & 58.03 & 58.58 & 56.63 & 58.14 \\
\bottomrule
\end{tabular}
\label{table:180K-ablations}
\end{table}

For each experiment, we generated 300–600 training samples of 180K tokens (concatenating two documents) using Qwen-2-72B and fine-tuned the data on LLaMA-3.1-8B-Instruct with a learning rate of 6e-5 for 1 epoch. As the 180K context length exceeds LLaMA-3.1-8B-Instruct’s native 128K window, we applied rope scaling. The following compositions were tested: a) \textbf{Random vs. fixed number of questions}: Follow-up questions were either randomized (2–10) or fixed (6 main and 4 follow-up) to maintain consistency. b) \textbf{Hierarchical vs. diverse and complex questions}: We tested hierarchical ordering questions (h) against questions targeting specific, diverse, and complex reasoning (s). Each experiment is labeled as x-y-z, where x refers to questions following the first document, y the second, and z to questions referencing the first document after the second is processed. For instance, h-h-s-fixed includes 6 hierarchical questions for each document and 4 diverse follow-ups referencing the first document after the second. c) \textbf{Summarization}: Some experiments excluded global summarization at the start to assess its impact on model comprehension.


Table \ref{table:180K-ablations} shows the ablation results on InfiniteBench. Notably: 1) All experiments outperformed the baseline LLaMA-3.1-8B-Instruct model by a significant margin, demonstrating the effectiveness of our strategy with rope scaling. 2) Fixed questions outperform randomized ones: hs-hs-hs-fixed scored 59.45, surpassing hs-hs-hs-randomized (58.51). 3) Hierarchical questions paired with diverse questions achieve the best performance: hs-hs-hs-fixed yielded the highest score (59.45), highlighting the benefits of structuring and diverse, complex questions. 4) Summarization improves performance: hs-hs-fixed-no-sum scored 58.03, slightly below hs-hs-hs-fixed (58.58). Based on these findings, for longer context lengths (Section \ref{sec:result-main}, we retain summarization, fix the number of questions/answers, and ensure both hierarchical and diverse questions are generated after direct documents and for those referencing previous ones.

\vspace{-3mm}
\section{Conclusion}
\label{sec:conclusion}
\vspace{-3mm}

This paper presents a novel strategy to generate high-quality, long-context instruction-tuning datasets that exceed the typical raw data context length. It incorporates hierarchical ordering to ensure logical coherence while maintaining diversity and complexity in questions. Systematic ablation studies show that combining diverse questions with hierarchical ordering enhances performance, particularly in long-context scenarios. Our 1M model demonstrates strong capabilities, outperforming LLaMA-3.1-8B-Instruct on InfiniteBench and significantly surpassing it on RULER, while maintaining robust performance on shorter-context tasks, as shown by LongBench and MMLU. 
Our data curation strategy is highly scalable, enabling efficient creation of instruction-tuning datasets exceeding 1 million tokens and scaling up to 10 million or more. With sufficient resources and a strong training stack, our method supports increasingly longer context lengths, potentially unlimited.


While our approach has significantly improved instruction tuning for long-context scenarios, a promising direction for future work is developing a self-evolutionary strategy that diversifies and adapts prompts. A short-context model could autonomously generate long-context instruction data using our methodology and evolve independently, creating diverse and adaptable prompts for various scenarios. This could enable models to progressively evolve into longer-context models. Additionally, combining our data-centric approach with architectural optimizations offers another promising avenue for future research.

\textbf{Ethics Statement.} In conducting this research, we ensured adherence to the highest ethical standards in the development and testing of our models. No human subjects were involved in data collection, ensuring that there are no privacy concerns or risks associated with the handling of personal information.

\textbf{Reproducibility.} We included the code to generate a bunch of hierarchical questions and diverse questions for a single document (see Section \ref{sec:coherent-single}) in supplementary material (see generating-data.py). We also included the code to concatenate multiple documents (see Section \ref{sec:multi-doc}) in supplementary material (see concatenate-350K.py). To enable long context training, we described detailed hardware setup in Section \ref{sec:setup}. Details about evaluations are also mentioned in in Section \ref{sec:setup}.

\label{others}

\bibliography{citations}

\begin{thebibliography}{30}
\providecommand{\natexlab}[1]{#1}
\providecommand{\url}[1]{\texttt{#1}}
\expandafter\ifx\csname urlstyle\endcsname\relax
  \providecommand{\doi}[1]{doi: #1}\else
  \providecommand{\doi}{doi: \begingroup \urlstyle{rm}\Url}\fi

\bibitem[Anil et~al.(2023)Anil, Dai, Firat, Johnson, Lepikhin, Passos, Shakeri, Taropa, Bailey, Chen, Chu, Clark, Shafey, Huang, Meier-Hellstern, Mishra, Moreira, Omernick, Robinson, Ruder, Tay, Xiao, Xu, Zhang, Abrego, Ahn, Austin, Barham, Botha, Bradbury, Brahma, Brooks, Catasta, Cheng, Cherry, Choquette-Choo, Chowdhery, Crepy, Dave, Dehghani, Dev, Devlin, Díaz, Du, Dyer, Feinberg, Feng, Fienber, Freitag, Garcia, Gehrmann, Gonzalez, Gur-Ari, Hand, Hashemi, Hou, Howland, Hu, Hui, Hurwitz, Isard, Ittycheriah, Jagielski, Jia, Kenealy, Krikun, Kudugunta, Lan, Lee, Lee, Li, Li, Li, Li, Li, Lim, Lin, Liu, Liu, Maggioni, Mahendru, Maynez, Misra, Moussalem, Nado, Nham, Ni, Nystrom, Parrish, Pellat, Polacek, Polozov, Pope, Qiao, Reif, Richter, Riley, Ros, Roy, Saeta, Samuel, Shelby, Slone, Smilkov, So, Sohn, Tokumine, Valter, Vasudevan, Vodrahalli, Wang, Wang, Wang, Wang, Wieting, Wu, Xu, Xu, Xue, Yin, Yu, Zhang, Zheng, Zheng, Zhou, Zhou, Petrov, and Wu]{anil2023palm2technicalreport}
Rohan Anil, Andrew~M. Dai, Orhan Firat, Melvin Johnson, Dmitry Lepikhin, Alexandre Passos, Siamak Shakeri, Emanuel Taropa, Paige Bailey, Zhifeng Chen, Eric Chu, Jonathan~H. Clark, Laurent~El Shafey, Yanping Huang, Kathy Meier-Hellstern, Gaurav Mishra, Erica Moreira, Mark Omernick, Kevin Robinson, Sebastian Ruder, Yi~Tay, Kefan Xiao, Yuanzhong Xu, Yujing Zhang, Gustavo~Hernandez Abrego, Junwhan Ahn, Jacob Austin, Paul Barham, Jan Botha, James Bradbury, Siddhartha Brahma, Kevin Brooks, Michele Catasta, Yong Cheng, Colin Cherry, Christopher~A. Choquette-Choo, Aakanksha Chowdhery, Clément Crepy, Shachi Dave, Mostafa Dehghani, Sunipa Dev, Jacob Devlin, Mark Díaz, Nan Du, Ethan Dyer, Vlad Feinberg, Fangxiaoyu Feng, Vlad Fienber, Markus Freitag, Xavier Garcia, Sebastian Gehrmann, Lucas Gonzalez, Guy Gur-Ari, Steven Hand, Hadi Hashemi, Le~Hou, Joshua Howland, Andrea Hu, Jeffrey Hui, Jeremy Hurwitz, Michael Isard, Abe Ittycheriah, Matthew Jagielski, Wenhao Jia, Kathleen Kenealy, Maxim Krikun, Sneha Kudugunta, Chang
  Lan, Katherine Lee, Benjamin Lee, Eric Li, Music Li, Wei Li, YaGuang Li, Jian Li, Hyeontaek Lim, Hanzhao Lin, Zhongtao Liu, Frederick Liu, Marcello Maggioni, Aroma Mahendru, Joshua Maynez, Vedant Misra, Maysam Moussalem, Zachary Nado, John Nham, Eric Ni, Andrew Nystrom, Alicia Parrish, Marie Pellat, Martin Polacek, Alex Polozov, Reiner Pope, Siyuan Qiao, Emily Reif, Bryan Richter, Parker Riley, Alex~Castro Ros, Aurko Roy, Brennan Saeta, Rajkumar Samuel, Renee Shelby, Ambrose Slone, Daniel Smilkov, David~R. So, Daniel Sohn, Simon Tokumine, Dasha Valter, Vijay Vasudevan, Kiran Vodrahalli, Xuezhi Wang, Pidong Wang, Zirui Wang, Tao Wang, John Wieting, Yuhuai Wu, Kelvin Xu, Yunhan Xu, Linting Xue, Pengcheng Yin, Jiahui Yu, Qiao Zhang, Steven Zheng, Ce~Zheng, Weikang Zhou, Denny Zhou, Slav Petrov, and Yonghui Wu.
\newblock Palm 2 technical report, 2023.
\newblock URL \url{https://arxiv.org/abs/2305.10403}.

\bibitem[Bai et~al.(2024)Bai, Lv, Zhang, Lyu, Tang, Huang, Du, Liu, Zeng, Hou, Dong, Tang, and Li]{bai2024longbenchbilingualmultitaskbenchmark}
Yushi Bai, Xin Lv, Jiajie Zhang, Hongchang Lyu, Jiankai Tang, Zhidian Huang, Zhengxiao Du, Xiao Liu, Aohan Zeng, Lei Hou, Yuxiao Dong, Jie Tang, and Juanzi Li.
\newblock Longbench: A bilingual, multitask benchmark for long context understanding, 2024.
\newblock URL \url{https://arxiv.org/abs/2308.14508}.

\bibitem[Beltagy et~al.(2020)Beltagy, Peters, and Cohan]{beltagy2020longformerlongdocumenttransformer}
Iz~Beltagy, Matthew~E. Peters, and Arman Cohan.
\newblock Longformer: The long-document transformer, 2020.
\newblock URL \url{https://arxiv.org/abs/2004.05150}.

\bibitem[Chen et~al.(2023)Chen, Wong, Chen, and Tian]{chen2023extendingcontextwindowlarge}
Shouyuan Chen, Sherman Wong, Liangjian Chen, and Yuandong Tian.
\newblock Extending context window of large language models via positional interpolation, 2023.
\newblock URL \url{https://arxiv.org/abs/2306.15595}.

\bibitem[Chen et~al.(2024)Chen, Qian, Tang, Lai, Liu, Han, and Jia]{chen2024longloraefficientfinetuninglongcontext}
Yukang Chen, Shengju Qian, Haotian Tang, Xin Lai, Zhijian Liu, Song Han, and Jiaya Jia.
\newblock Longlora: Efficient fine-tuning of long-context large language models, 2024.
\newblock URL \url{https://arxiv.org/abs/2309.12307}.

\bibitem[Gu \& Dao(2024)Gu and Dao]{gu2024mambalineartimesequencemodeling}
Albert Gu and Tri Dao.
\newblock Mamba: Linear-time sequence modeling with selective state spaces, 2024.
\newblock URL \url{https://arxiv.org/abs/2312.00752}.

\bibitem[Guo et~al.(2022)Guo, Ainslie, Uthus, Ontanon, Ni, Sung, and Yang]{guo2022longt5efficienttexttotexttransformer}
Mandy Guo, Joshua Ainslie, David Uthus, Santiago Ontanon, Jianmo Ni, Yun-Hsuan Sung, and Yinfei Yang.
\newblock Longt5: Efficient text-to-text transformer for long sequences, 2022.
\newblock URL \url{https://arxiv.org/abs/2112.07916}.

\bibitem[Hendrycks et~al.(2021)Hendrycks, Burns, Basart, Zou, Mazeika, Song, and Steinhardt]{hendrycks2021measuringmassivemultitasklanguage}
Dan Hendrycks, Collin Burns, Steven Basart, Andy Zou, Mantas Mazeika, Dawn Song, and Jacob Steinhardt.
\newblock Measuring massive multitask language understanding, 2021.
\newblock URL \url{https://arxiv.org/abs/2009.03300}.

\bibitem[Ho et~al.(2024)Ho, Bae, Kim, Jo, Kim, Schuster, Fisch, Thorne, and Yun]{ho2024blocktransformerglobaltolocallanguage}
Namgyu Ho, Sangmin Bae, Taehyeon Kim, Hyunjik Jo, Yireun Kim, Tal Schuster, Adam Fisch, James Thorne, and Se-Young Yun.
\newblock Block transformer: Global-to-local language modeling for fast inference, 2024.
\newblock URL \url{https://arxiv.org/abs/2406.02657}.

\bibitem[Hsieh et~al.(2024)Hsieh, Sun, Kriman, Acharya, Rekesh, Jia, Zhang, and Ginsburg]{hsieh2024rulerwhatsrealcontext}
Cheng-Ping Hsieh, Simeng Sun, Samuel Kriman, Shantanu Acharya, Dima Rekesh, Fei Jia, Yang Zhang, and Boris Ginsburg.
\newblock Ruler: What's the real context size of your long-context language models?, 2024.
\newblock URL \url{https://arxiv.org/abs/2404.06654}.

\bibitem[Jiang et~al.(2023)Jiang, Sablayrolles, Mensch, Bamford, Chaplot, de~las Casas, Bressand, Lengyel, Lample, Saulnier, Lavaud, Lachaux, Stock, Scao, Lavril, Wang, Lacroix, and Sayed]{jiang2023mistral7b}
Albert~Q. Jiang, Alexandre Sablayrolles, Arthur Mensch, Chris Bamford, Devendra~Singh Chaplot, Diego de~las Casas, Florian Bressand, Gianna Lengyel, Guillaume Lample, Lucile Saulnier, Lélio~Renard Lavaud, Marie-Anne Lachaux, Pierre Stock, Teven~Le Scao, Thibaut Lavril, Thomas Wang, Timothée Lacroix, and William~El Sayed.
\newblock Mistral 7b, 2023.
\newblock URL \url{https://arxiv.org/abs/2310.06825}.

\bibitem[Katharopoulos et~al.(2020)Katharopoulos, Vyas, Pappas, and Fleuret]{katharopoulos2020transformersrnnsfastautoregressive}
Angelos Katharopoulos, Apoorv Vyas, Nikolaos Pappas, and François Fleuret.
\newblock Transformers are rnns: Fast autoregressive transformers with linear attention, 2020.
\newblock URL \url{https://arxiv.org/abs/2006.16236}.

\bibitem[Li et~al.(2024)Li, Wang, Zheng, and Zhang]{li2024looglelongcontextlanguagemodels}
Jiaqi Li, Mengmeng Wang, Zilong Zheng, and Muhan Zhang.
\newblock Loogle: Can long-context language models understand long contexts?, 2024.
\newblock URL \url{https://arxiv.org/abs/2311.04939}.

\bibitem[Liu et~al.(2023)Liu, Lin, Hewitt, Paranjape, Bevilacqua, Petroni, and Liang]{liu2023lostmiddlelanguagemodels}
Nelson~F. Liu, Kevin Lin, John Hewitt, Ashwin Paranjape, Michele Bevilacqua, Fabio Petroni, and Percy Liang.
\newblock Lost in the middle: How language models use long contexts, 2023.
\newblock URL \url{https://arxiv.org/abs/2307.03172}.

\bibitem[Men et~al.(2024)Men, Xu, Wang, Zhang, Lin, Han, and Chen]{men2024baseropeboundscontext}
Xin Men, Mingyu Xu, Bingning Wang, Qingyu Zhang, Hongyu Lin, Xianpei Han, and Weipeng Chen.
\newblock Base of rope bounds context length, 2024.
\newblock URL \url{https://arxiv.org/abs/2405.14591}.

\bibitem[Meng et~al.(2023)Meng, Bau, Andonian, and Belinkov]{meng2023locatingeditingfactualassociations}
Kevin Meng, David Bau, Alex Andonian, and Yonatan Belinkov.
\newblock Locating and editing factual associations in gpt, 2023.
\newblock URL \url{https://arxiv.org/abs/2202.05262}.

\bibitem[Munkhdalai et~al.(2024)Munkhdalai, Faruqui, and Gopal]{munkhdalai2024leavecontextbehindefficient}
Tsendsuren Munkhdalai, Manaal Faruqui, and Siddharth Gopal.
\newblock Leave no context behind: Efficient infinite context transformers with infini-attention, 2024.
\newblock URL \url{https://arxiv.org/abs/2404.07143}.

\bibitem[Peng et~al.(2023)Peng, Quesnelle, Fan, and Shippole]{peng2023yarnefficientcontextwindow}
Bowen Peng, Jeffrey Quesnelle, Honglu Fan, and Enrico Shippole.
\newblock Yarn: Efficient context window extension of large language models, 2023.
\newblock URL \url{https://arxiv.org/abs/2309.00071}.

\bibitem[Press et~al.(2022)Press, Smith, and Lewis]{press2022trainshorttestlong}
Ofir Press, Noah~A. Smith, and Mike Lewis.
\newblock Train short, test long: Attention with linear biases enables input length extrapolation, 2022.
\newblock URL \url{https://arxiv.org/abs/2108.12409}.

\bibitem[Su et~al.(2023)Su, Lu, Pan, Murtadha, Wen, and Liu]{su2023roformerenhancedtransformerrotary}
Jianlin Su, Yu~Lu, Shengfeng Pan, Ahmed Murtadha, Bo~Wen, and Yunfeng Liu.
\newblock Roformer: Enhanced transformer with rotary position embedding, 2023.
\newblock URL \url{https://arxiv.org/abs/2104.09864}.

\bibitem[Sun et~al.(2022)Sun, Dong, Patra, Ma, Huang, Benhaim, Chaudhary, Song, and Wei]{sun2022lengthextrapolatabletransformer}
Yutao Sun, Li~Dong, Barun Patra, Shuming Ma, Shaohan Huang, Alon Benhaim, Vishrav Chaudhary, Xia Song, and Furu Wei.
\newblock A length-extrapolatable transformer, 2022.
\newblock URL \url{https://arxiv.org/abs/2212.10554}.

\bibitem[Touvron et~al.(2023)Touvron, Lavril, Izacard, Martinet, Lachaux, Lacroix, Rozière, Goyal, Hambro, Azhar, Rodriguez, Joulin, Grave, and Lample]{touvron2023llamaopenefficientfoundation}
Hugo Touvron, Thibaut Lavril, Gautier Izacard, Xavier Martinet, Marie-Anne Lachaux, Timothée Lacroix, Baptiste Rozière, Naman Goyal, Eric Hambro, Faisal Azhar, Aurelien Rodriguez, Armand Joulin, Edouard Grave, and Guillaume Lample.
\newblock Llama: Open and efficient foundation language models, 2023.
\newblock URL \url{https://arxiv.org/abs/2302.13971}.

\bibitem[Wang et~al.(2023)Wang, Kordi, Mishra, Liu, Smith, Khashabi, and Hajishirzi]{wang2023selfinstructaligninglanguagemodels}
Yizhong Wang, Yeganeh Kordi, Swaroop Mishra, Alisa Liu, Noah~A. Smith, Daniel Khashabi, and Hannaneh Hajishirzi.
\newblock Self-instruct: Aligning language models with self-generated instructions, 2023.
\newblock URL \url{https://arxiv.org/abs/2212.10560}.

\bibitem[Wei et~al.(2022)Wei, Tay, Bommasani, Raffel, Zoph, Borgeaud, Yogatama, Bosma, Zhou, Metzler, Chi, Hashimoto, Vinyals, Liang, Dean, and Fedus]{wei2022emergentabilitieslargelanguage}
Jason Wei, Yi~Tay, Rishi Bommasani, Colin Raffel, Barret Zoph, Sebastian Borgeaud, Dani Yogatama, Maarten Bosma, Denny Zhou, Donald Metzler, Ed~H. Chi, Tatsunori Hashimoto, Oriol Vinyals, Percy Liang, Jeff Dean, and William Fedus.
\newblock Emergent abilities of large language models, 2022.
\newblock URL \url{https://arxiv.org/abs/2206.07682}.

\bibitem[Wu et~al.(2023)Wu, Irsoy, Lu, Dabravolski, Dredze, Gehrmann, Kambadur, Rosenberg, and Mann]{wu2023bloomberggptlargelanguagemodel}
Shijie Wu, Ozan Irsoy, Steven Lu, Vadim Dabravolski, Mark Dredze, Sebastian Gehrmann, Prabhanjan Kambadur, David Rosenberg, and Gideon Mann.
\newblock Bloomberggpt: A large language model for finance, 2023.
\newblock URL \url{https://arxiv.org/abs/2303.17564}.

\bibitem[Xiong et~al.(2023)Xiong, Liu, Molybog, Zhang, Bhargava, Hou, Martin, Rungta, Sankararaman, Oguz, Khabsa, Fang, Mehdad, Narang, Malik, Fan, Bhosale, Edunov, Lewis, Wang, and Ma]{xiong2023effectivelongcontextscalingfoundation}
Wenhan Xiong, Jingyu Liu, Igor Molybog, Hejia Zhang, Prajjwal Bhargava, Rui Hou, Louis Martin, Rashi Rungta, Karthik~Abinav Sankararaman, Barlas Oguz, Madian Khabsa, Han Fang, Yashar Mehdad, Sharan Narang, Kshitiz Malik, Angela Fan, Shruti Bhosale, Sergey Edunov, Mike Lewis, Sinong Wang, and Hao Ma.
\newblock Effective long-context scaling of foundation models, 2023.
\newblock URL \url{https://arxiv.org/abs/2309.16039}.

\bibitem[Xu et~al.(2023)Xu, Sun, Zheng, Geng, Zhao, Feng, Tao, and Jiang]{xu2023wizardlmempoweringlargelanguage}
Can Xu, Qingfeng Sun, Kai Zheng, Xiubo Geng, Pu~Zhao, Jiazhan Feng, Chongyang Tao, and Daxin Jiang.
\newblock Wizardlm: Empowering large language models to follow complex instructions, 2023.
\newblock URL \url{https://arxiv.org/abs/2304.12244}.

\bibitem[Zeng et~al.(2024)Zeng, Xu, Zhao, Lou, and Chen]{zeng2024automaticinstructionevolvinglarge}
Weihao Zeng, Can Xu, Yingxiu Zhao, Jian-Guang Lou, and Weizhu Chen.
\newblock Automatic instruction evolving for large language models, 2024.
\newblock URL \url{https://arxiv.org/abs/2406.00770}.

\bibitem[Zhang et~al.(2024)Zhang, Chen, Hu, Xu, Chen, Hao, Han, Thai, Wang, Liu, and Sun]{zhang2024inftybenchextendinglongcontext}
Xinrong Zhang, Yingfa Chen, Shengding Hu, Zihang Xu, Junhao Chen, Moo~Khai Hao, Xu~Han, Zhen~Leng Thai, Shuo Wang, Zhiyuan Liu, and Maosong Sun.
\newblock $\infty$bench: Extending long context evaluation beyond 100k tokens, 2024.
\newblock URL \url{https://arxiv.org/abs/2402.13718}.

\bibitem[Zhao et~al.(2024)Zhao, Wei, Zeng, Cheng, Yang, Cheng, Wang, Li, Wu, Zhu, Gan, Hu, Yan, Fang, and Zhou]{zhao2024longskyworktrainingrecipeefficiently}
Liang Zhao, Tianwen Wei, Liang Zeng, Cheng Cheng, Liu Yang, Peng Cheng, Lijie Wang, Chenxia Li, Xuejie Wu, Bo~Zhu, Yimeng Gan, Rui Hu, Shuicheng Yan, Han Fang, and Yahui Zhou.
\newblock Longskywork: A training recipe for efficiently extending context length in large language models, 2024.
\newblock URL \url{https://arxiv.org/abs/2406.00605}.

\end{thebibliography}
\bibliographystyle{iclr2025_conference}

\newpage
\appendix

\section{Appendix: Additional Details on Data Generation Algorithms}
In this section, we present the pseudocode for the hierarchical QA generation strategy described in Section \ref{sec:coherent-single}, along with the algorithm for combining multiple documents, as outlined in Section \ref{sec:multi-doc}.
\label{sec:additional-details-algorithms}

\begin{algorithm}
\caption{Hierarchical Question Generation Strategy (Single Document)}
\begin{algorithmic}[1]
\Procedure{GenerateExtendedContext}{document, N\_Questions\_To\_Generate}
    \State chunks $\gets$ HierarchicalSplit(document.text)
    \State summaries, full\_summary $\gets$ SummarizeHierarchical(chunks)
    \State conversations $\gets$ [InitialSummary(document.text, full\_summary)]
    \For{$i = 1$ to N\_Questions\_To\_Generate}
        \State context, summary $\gets$ SelectContext(chunks, summaries, last\_medium, last\_small, $i$)
        \State qa\_pair $\gets$ GenerateQAPair(context, summary)
        \State AppendToConversations(conversations, qa\_pair)
        \State UpdateLastChunks(last\_medium, last\_small)
    \EndFor
    \State \Return conversations
\EndProcedure

\Procedure{SelectContext}{chunks, summaries, last\_medium, last\_small, iteration\_index}
    \If{first iteration}
        \State \Return random medium chunk
    \ElsIf{no small chunk selected}
        \State \Return first small chunk of current medium
    \Else
        \State random\_choice $\gets$ RandomChoice([0, 1, 2])  \Comment{Equal 1/3 probability for each}
        \If{random\_choice = 0}
            \State \Return deeper content of current small chunk
        \ElsIf{random\_choice = 1}
            \State \Return next small chunk in current medium
        \Else
            \State \Return next medium chunk
        \EndIf
    \EndIf
\EndProcedure

\Procedure{GenerateQAPair}{context, summary}
    \If{ContextIsSpecific(context)}
        \State \Return GenerateSpecificQAPair(context)
    \Else
        \State \Return GenerateGeneralQAPair(context, summary)
    \EndIf
\EndProcedure
\end{algorithmic}
\end{algorithm}

\begin{algorithm}
\caption{Concatenating Multiple Documents}
\begin{algorithmic}

\State \textbf{Input:} Set of $K$ documents, each with hierarchical and diverse questions
\State \textbf{Initialize:} conversation list $C \leftarrow \emptyset$

\For{each document $D_i$ where $i = 1, 2, \dots, K$}
    \State $H_i \leftarrow \text{GenerateHierarchicalQuestions}(D_i)$
    \State $S_i \leftarrow \text{RandomlySampleSpecificQuestions}(D_i)$
    \State $C \leftarrow C \cup \text{InitialHierarchicalQuestions}(H_i)$
    \State $C \leftarrow C \cup \text{RandomlySampleDiverseQuestions}(S_i)$
    \State Store remaining unselected diverse questions from $S_i$
\EndFor

\For{each document $D_i$ where $i = 2, 3, \dots, K$}
    \State $C \leftarrow C \cup \text{NextHierarchicalQuestions}(H_{i-1})$
    \State $C \leftarrow C \cup \text{RandomlySampleUnselectedDiverse}(S_{i-1})$
    \State Update hierarchical index for document $D_i$
\EndFor

\For{each document $D_i$ where $i = 1, 2, \dots, K-1$}
    \If{RandomCondition$(0.6)$}
        \State $C \leftarrow C \cup \text{FollowUpHierarchicalQuestions}(H_i)$
    \EndIf
\EndFor

\State Process remaining specific and diverse questions:
\State $x \leftarrow \frac{\text{Length}(S_i)}{2}$ 
\If{$x \geq \text{ThresholdForSpecificQuestions}$}
    \State Select and append follow-up specific questions to $C$
    \State Remove selected follow-up specific questions from pool
\EndIf

\State \textbf{Output:} Final conversation list $C$

\end{algorithmic}
\end{algorithm}

\section{Additional Information on Data Generation Prompts}
\label{apx:prompts}
Here we list all prompts used in the different stages of our synthetic data generation pipeline.

\begin{tcolorbox}[colback=blue!5!white,colframe=blue!75!black,title=Document Summarization]
\begin{verbatim}
"""Summarize the following text concisely in no
    more than {word_limit} words:

    {chunk}"""
\end{verbatim}
\end{tcolorbox}

\begin{tcolorbox}[colback=blue!5!white,colframe=blue!75!black,title=Diverse Questions]
\begin{verbatim}
    """Context information is below.
    ---------------------
    ${context}
    ---------------------
    Given the context information and not prior knowledge.
    Generate content based on the below query.
    You are a Teacher/Professor. Your task is to
    set up 1 diverse temporal question about the
    context for an upcoming quiz/examination. The question
    should cover different time periods and events
    described in the context. Restrict the question
    to the context information provided. You must
    return the result in JSON: {'question': <question>,
    'answer': <answer>}"""
\end{verbatim}
\end{tcolorbox}

\begin{tcolorbox}[colback=blue!5!white,colframe=blue!75!black,title=Diverse Questions]
\begin{verbatim}
    """Context information is below.
    ---------------------
    ${context}
    ---------------------
    Given the context information and not prior knowledge.
    Generate content based on the below query.
    You are a Teacher/Professor. Your task is to
    create 1 character-based question from the context
    for an upcoming quiz/examination. The question should
    explore different aspects of the characters, such
    as their motivations, actions, and relationships. Restrict
    the question to the context information provided.
    You must return the result in JSON:
    {'question': <question>, 'answer': <answer>}"""
    
    """Context information is below.
    ---------------------
    ${context}
    ---------------------
    Given the context information and not prior knowledge.
    Generate content based on the below query.
    Formulate 1 complex question that requires analysis
    of multiple aspects from the context for
    an upcoming quiz/examination. The question should encourage
    critical thinking and synthesis of different pieces
    of information within the context. Restrict the
    question to the context information provided. You
    must return the result in JSON: {'question':
    <question>, 'answer': <answer>}"""
    
    """Context information is below.
    ---------------------
    ${context}
    ---------------------
    Given the context information and not prior knowledge.
    Generate content based on the below query.
    You are a Teacher/Professor. Ask 1 question
    about the main themes or messages of
    the text for an upcoming quiz/examination. The
    question should cover different aspects of the
    themes and how they are developed in
    the context. Restrict the question to the
    context information provided. You must return the
    result in JSON: {'question': <question>, 
    'answer': <answer>}"""

    
\end{verbatim}
\end{tcolorbox}

\begin{tcolorbox}[colback=blue!5!white,colframe=blue!75!black,title=Diverse Questions]
\begin{verbatim}
    """Context information is below.
    ---------------------
    ${context}
    ---------------------
    Given the context information and not prior knowledge.
    Generate content based on the below query.
    You are a Teacher/Professor. Create 1 question
    that compare different elements within the context
    for an upcoming quiz/examination. The question should
    highlight similarities and differences between various 
    elements such as characters, events, and themes. Restrict
    the question to the context information provided.
    You must return the result in JSON:
    {'question': <question>, 'answer': <answer>}"""

    """Context information is below.
    ---------------------
    ${context}
    ---------------------
    Given the context information and not prior knowledge.
    Generate content based on the below query.
    You are a Teacher/Professor. Develop 1 question
    that explore the cause and effect relationships
    within the context for an upcoming quiz/examination.
    The question should focus on understanding the
    reasons behind events and their outcomes. Restrict
    the question to the context information provided.
    You must return the result in JSON:
    {'question': <question>, 'answer': <answer>}"""
    
    """Context information is below.
    ---------------------
    ${context}
    ---------------------
    Given the context information and not prior knowledge.
    Generate content based on the below query.
    You are a Teacher/Professor. Create 1 hypothetical
    question based on the context for an
    upcoming quiz/examination. The question should explore 
    what-if scenarios and possible alternate outcomes. Restrict 
    the question to the context information provided. You
    must return the result in JSON: {'question':
    <question>, 'answer': <answer>}"""
    
   

\end{verbatim}
\end{tcolorbox}

\begin{tcolorbox}[colback=blue!5!white,colframe=blue!75!black,title=Diverse Questions]
\begin{verbatim}
 """Context information is below.
    ---------------------
    ${context}
    ---------------------
    Given the context information and not prior knowledge.
    Generate content based on the below query.
    You are a Teacher/Professor. Formulate 1 question
    that require interpretation of the context for
    an upcoming quiz/examination. The question should encourage
    students to provide their own insights and
    interpretations based on the information given. Restrict
    the question to the context information provided.
    You must return the result in JSON:
    {'question': <question>, 'answer': <answer>}"""

    """Context information is below.
    ---------------------
    ${context}
    ---------------------
    Given the context information and not prior knowledge.
    Generate content based on the below query.
    You are a Teacher/Professor. Ask 1 detail-oriented
    question about the context for an upcoming
    quiz/examination. These question should focus on specific
    details, facts, and figures mentioned in the
    context. Restrict the question to the context
    information provided. You must return the result
    in JSON: {'question': <question>, 'answer': <answer>}"""

    """Context information is below.
    ---------------------
    ${context}
    ---------------------
    Given the context information and not prior knowledge.
    Generate content based on the below query.
    You are a Teacher/Professor. Create 1 question
    that explore different perspectives or viewpoints within
    the context for an upcoming quiz/examination. The
    question should examine how different characters or
    groups might view events or themes differently.
    Restrict the questions to the context information
    provided. You must return the result in
    JSON: {'question': <question>, 'answer': <answer>}"""
\end{verbatim}
\end{tcolorbox}

\begin{tcolorbox}[colback=purple!5!white,colframe=blue!75!black,title=Multi-Hop Questions]
\begin{verbatim}
"""Context information is below.
${selected_chunk_1}
${selected_chunk_2}
${selected_chunk_3}
You are a Professor designing a final exam
for an advanced interdisciplinary course. Create 1
complex question requiring deep analysis and synthesis of
information from all three chunks. Do not mention
that there are three chunks/your questions. Do not
mention excerpts either. For example, instead of a
question that says "Analyze the theme of justice
and its various forms as portrayed in the
three provided literary excerpts. How do the characters'
actions and the outcomes of their situations reflect
or challenge traditional notions of justice? Consider the
legal, personal, and societal implications of justice in
each excerpt and discuss the role of power
dynamics in shaping justice." You should say: "Analyze
the theme of justice and its various forms
as portrayed. How do the characters' actions and
the outcomes of their situations reflect or challenge
traditional notions of justice? Consider the legal, personal,
and societal implications of justice and discuss the
role of power dynamics in shaping justice."
Question Guidelines:
1. The question must integrate and require reasoning
across all three chunks.
2. The question should be multi-layered, promoting analysis,
synthesis, and evaluation.
Answer Guidelines:
1. Provide a comprehensive answer addressing all question
aspects.
2. Reference and interconnect information from each chunk.
Return 1 question-answer pair in JSON format:
{ "question": <question>, "answer": <answer> }
"""
\end{verbatim}
\end{tcolorbox}

\begin{tcolorbox}[colback=green!5!white,colframe=green!75!black,title=Specific Detail Question]
\begin{verbatim}
"""Context information is below.
${context}
Given the context information and not prior knowledge,
generate content based on the below query. You
are a Teacher/Professor. Create 1 specific question about
the details, events, characters, and settings from the
context provided. This question should have an exact,
unambiguous answer that can be directly found in
the given information. The question should be similar
in style to the following examples:
"Where does Character A meet Character B for
the first time?"
"What is Character C's religion?"
"Where does Character D live for the majority
of the story?"
"Which of the following is NOT one of
Character E's responsibilities?"
"Which among [list of names] is not [Character
F]'s child?"
"Who among [list of names] is the final
to perish?"
"What's the name of [Family name]'s summer home?"
"Who accompanied [Character G] [specific activity] at last?"
Ensure that the question and answer are strictly
based on the context information provided. The question
may include multiple-choice options when appropriate. You
must return the result in JSON: {'question': <question>,
'answer': <answer>}"""
\end{verbatim}
\end{tcolorbox}

\section{Complete Conversation Data}
\label{sec:additional-data-example}

Here we present an example of our instruction tuning data that is developed by concatenating four documents (350K context length). The data consists of three types of content: long documents and their summarizations (yellow) (constituting QA pairs), QA pairs extracted from each document (blue), and QA pairs that reference past documents (green). These are color-coded in the following example for clarity.

\begin{lstlisting}[style=yellowstyle]
{"conversations": [
  {
    "role": "user",
    "content": "Produced by Charles Bowen from page scans provided by the Web Archive\n\n\n\n\nTranscriber's Notes: ... Please give me a summary of the book"
  },
  {
    "role": "assistant",
    "content": "Charles Bowen created a digital book from Web Archive scans, ..."
  },
\end{lstlisting}

\begin{lstlisting}[style=bluestyle]
  {
    "role": "user",
    "content": "What are the main reasons for the tensions and conflicts among the characters in 'The Patricians'?"
  },
  {
    "role": "assistant",
    "content": "The tensions and conflicts in 'The Patricians' arise from issues of pride, ..."
  },
  {
    "role": "user",
    "content": "What class in ancient Rome did the term 'Patricians' typically refer to?"
  },
  {
    "role": "assistant",
    "content": "The term 'Patricians' typically referred to the aristocratic class in ancient Rome, who were affluent, powerful, and held considerable political sway in the Roman Republic."
  },
  {
    "role": "user",
    "content": "Who among the following did not attend Christopher's banquet? A) Tausdorf B) Althea C) Schindel D) Francis"
  },
  {
    "role": "assistant",
    "content": "D) Francis"
  },
  {
    "role": "user",
    "content": "Analyze the theme of justice and its various forms as portrayed. How do the characters' actions and the outcomes of their situations reflect or challenge traditional notions of justice? Consider the legal, personal, and societal implications of justice and discuss the role of power dynamics in shaping justice."
  },
  {
    "role": "assistant",
    "content": "The theme of justice is explored through the complex interplay of legal, personal, and societal implications, reflecting and challenging traditional notions of justice. ..."
  },
  ...
\end{lstlisting}
\begin{lstlisting}[style=yellowstyle]
  {
    "role": "user",
    "content": "Produced by D Alexander and the Online Distributed Proofreading Team at http://www.pgdp.net ... + Please give me a summary of the book"
  },
  {
    "role": "assistant",
    "content": "The text acknowledges D Alexander ..."
  },
\end{lstlisting}

\begin{lstlisting}[style=bluestyle]
  {
    "role": "user",
    "content": "Who does the professor and his party meet at the sulphur spring?"
  },
  {
    "role": "assistant",
    "content": "Jranvin, the chief of the islanders"
  },
  {
    "role": "user",
    "content": "What does Jim recognize that lifts his spirits while confined 
    on the Marjorie?"
  },
  {
    "role": "assistant",
    "content": "Jim recognizes his own ship, which lifts his spirits while he is 
    confined on the Marjorie."
  },
  {
    "role": "user",
    "content": "How did the professor and his party find out about the location 
    of the Marjorie and what did they discover upon investigating the harbor?"
  },
  {
    "role": "assistant",
    "content": "The professor and his party, including Tom, Jo, and Berwick, 
    hiked to a cliff and climbed to the top where they spotted the Marjorie 
    and another ship in a harbor five miles away. Upon investigating the harbor, 
    they found the Sea Eagle leaving and chaos on the Marjorie. They also captured 
    the steward who revealed he helped Jim escape."
  },
  ...
  \end{lstlisting}
  \begin{lstlisting}[style=greenstyle]
  {
    "role": "user",
    "content": "Analyze the theme of justice and its various forms as portrayed.
    How do the characters' actions and the outcomes of their situations reflect 
    or challenge traditional notions of justice? Consider the legal, personal,
    and societal implications of justice and discuss the role of power dynamics 
    in shaping justice."
  },
  {
    "role": "assistant",
    "content": "The theme of justice is intricately woven through 
    the narrative, ..."
  },
  ...
  \end{lstlisting}
  \begin{lstlisting}[style=yellowstyle]
  {
    "role": "user",
    "content": "\n\n\n\nProduced by Marius Masi, Juliet Sutherland and the Online
    \nDistributed Proofreading Team at ...  Please give me a summary of the book"
  },
  {
    "role": "assistant",
    "content": "The text describes a collection of illustrated books about Japan, ..."
  },
  \end{lstlisting}
  \begin{lstlisting}[style=bluestyle]
  {
    "role": "user",
    "content": "How does the text compare Japanese and Western art in terms 
    of their integration into daily life and appreciation by the public?"
  },
  {
    "role": "assistant",
    "content": "The text emphasizes that Japanese art, unlike Western art, 
    is deeply integrated into daily life and universally appreciated,
    similar to the vitality of art in ancient Greece..."
  },
  {
    "role": "user",
    "content": "How does the book contrast Japanese and Western art in terms of
    their integration into daily life and societal appreciation?"
  },
  {
    "role": "assistant",
    "content": "The book highlights that Japanese art is more integrated into daily
    life and universally appreciated, ..."
  },
  ...
  \end{lstlisting}
  \begin{lstlisting}[style=greenstyle]
  {
    "role": "user",
    "content": "Analyze the character development of Tausdorf throughout the
    passage. How does his perception of Bona change, and what factors contribute 
    to this change? Additionally, discuss the role of the setting and the use of 
    imagery in shaping Tausdorf's actions and emotions."
  },
  {
    "role": "assistant",
    "content": "Tausdorf's character development is complex and multifaceted, ..."
  },
  {
    "role": "user",
    "content": "What is the name of the yacht that Jim and Berwick consider hiring?"
  },
  {
    "role": "assistant",
    "content": "Storm King"
  },
  ...
  \end{lstlisting}
  \begin{lstlisting}[style=yellowstyle]
  {"role": "user", "content": "\n\n\n\nProduced by David Edwards and the Online Distributed ... Please give me a summary of the book"},
  {"role": "assistant", "content": "\"The Curlytops Snowed In\" by Howard R. Garis follows siblings Ted and Janet Martin, known as the Curlytops, ..."},
  \end{lstlisting}
  \begin{lstlisting}[style=bluestyle]
  {"role": "user", "content": "What challenge did Nicknack, the goat, face during the winter adventures, and how did the children address it?"}, 
  {"role": "assistant", "content": "Nicknack had difficulty standing on ice due to the unsuitability of his hooves..."}, 
  {"role": "user", "content": "What causes the snow house to collapse in Chapter VI, and what do the children decide to do after the incident?"}, 
  {"role": "assistant", "content": "The snow house collapses when Nicknack the goat ..."},
  ...
  \end{lstlisting}
  \begin{lstlisting}[style=greenstyle]
  {"role": "user", "content": "Compare and contrast the portrayal of art and its integration into daily life in Japan and the West, as described in the text. How does the Japanese approach to art, particularly in the context of theater, reflect broader cultural values and philosophies? Discuss the role of realism, balance, and the environment in shaping the artistic experience in both cultures."}, 
  {"role": "assistant", "content": "The text illustrates a stark contrast between the Japanese and Western approaches to art, particularly in theater, which reflects deeper cultural values and philosophies. In Japan, art is not merely a conscious effort to achieve beauty but an instinctive expression of inherited taste that permeates all aspects of life..."},
  {"role": "user", "content": "Analyze the theme of survival and its various forms as portrayed in the narratives. How do the characters' actions and the outcomes of their situations reflect or challenge traditional notions of survival? Consider the natural, social, and personal implications of survival and discuss the role of resourcefulness in shaping survival strategies."}, 
  {"role": "assistant", "content": "The theme of survival is intricately woven through the narratives, reflecting the characters' resilience and adaptability in the ..."},
  ...
]}
\end{lstlisting}


\section{RULER numerical results}

\label{sec:ruler-numerical}

\begin{table}[t]
\centering
\small
\setlength{\tabcolsep}{4pt}
\begin{tabular}{@{}ccc@{}}
\begin{tabular}[t]{@{}r@{\hspace{0.5em}}l@{}}
\toprule
\multicolumn{2}{c}{\textbf{1M model}} \\
\textbf{Context Length} & \textbf{Accuracy (\%)} \\
\midrule
1 000 000 & 62.95\% \\
524 288 & 76.88\% \\
262 144 & 76.24 \% \\
131 072 & 82.05 \% \\
65 536 & 83.73 \% \\
32 768 & 86.19 \% \\
16 384 & 90.06 \% \\
8 192 & 92.96 \% \\
4 096 & 95.59 \%  \\
\bottomrule
\end{tabular}
&
\begin{tabular}[t]{@{}r@{\hspace{0.5em}}l@{}}
\toprule
\multicolumn{2}{c}{\textbf{650K model}} \\
\textbf{Context Length} & \textbf{Accuracy (\%)} \\
\midrule
650 000 & 57.24 \%\\
524 288 & 74.88 \%\\
262 144 & 78.26 \%\\
131 072 & 82.29 \%\\
65 536 & 81.81 \%\\
32 768 & 87.96 \%\\
16 384 & 92.86 \%\\
8 192 & 91.92 \%\\
4 096 & 95.49 \%\\
\bottomrule
\end{tabular}
&
\begin{tabular}[t]{@{}r@{\hspace{0.5em}}l@{}}
\toprule
\multicolumn{2}{c}{\textbf{350K model}} \\
\textbf{Context Length} & \textbf{Accuracy (\%)} \\
\midrule
350 000 & 71.15 \% \\
262 144 & 78.26 \% \\
131 072 & 82.72 \% \\
65 536 & 82.63 \% \\
32 768 & 87.26 \% \\
16 384 & 91.45 \% \\
8 192 & 92.62 \% \\
4 096 & 96.13 \% \\
\bottomrule
\end{tabular}
\end{tabular}
\caption{Context length accuracy of RULER for different models}
\label{tab:ruler_data_three_models}
\end{table}

\begin{table}[t]
\centering
\small
\caption{Context length of RULER for LLaMA-3.1-8B-Instruct models}
\label{tab:ruler_llama_comparison}
\setlength{\tabcolsep}{6pt}
\begin{tabular}{@{}cc@{}}
\begin{tabular}[t]{@{}r@{\hspace{0.5em}}l@{}}
\toprule
\multicolumn{2}{c}{\textbf{LLaMA-3.1-8B-Instruct}} \\
\textbf{Context Length} & \textbf{Percentage (\%)} \\
\midrule  
524 288 & 28.93 \% \\
262 144 & 30.34 \% \\
131 072 & 83.06 \% \\
65 536 & 82.26 \% \\
32 768 & 88.44 \% \\
16 384 & 93.13 \% \\
8 192 & 92.08 \% \\
4 096 & 95.49 \% \\
\bottomrule
\end{tabular}
&
\begin{tabular}[t]{@{}r@{\hspace{0.5em}}l@{}}
\toprule
\multicolumn{2}{c}{\textbf{Zero-shot Rope Scaling to 1M}} \\
\textbf{Context Length} & \textbf{Percentage (\%)} \\
\midrule
1 000 000 & 48.81 \% \\
524 288 & 62.53 \% \\
262 144 &  66.44 \% \\
131 072 & 72.68 \% \\
65 536 & 77.81 \% \\
32 768 & 84.01 \% \\
16 384 & 87.36 \% \\
8 192 & 90.73 \% \\
4 096 & 95.94 \% \\
\bottomrule
\end{tabular} \\
\\
\small{(a) Context length of RULER} & \small{(b) Context length of RULER with} \\
\small{of LLaMA-3.1-8B-Instruct} & \small{zero-shot rope scaling to 1M context length}
\end{tabular}
\end{table}

\section{100K Context Length Ablation Studies}

\label{sec:ablation-100K}
The 100K ablation studies aim to assess whether hierarchical ordering and diverse question types improve results on single-document instruction tuning data.
We also aim to identify which of these factors most significantly influences overall performance.  In particular, we want to explore (1) whether hierarchical ordering enhances outcomes, (2) whether diverse question sets contribute positively, and (3) whether the use of multi-hop questions further boosts results.

Each experiment uses 300-600 data samples, each with 100K tokens, fine-tuned on LLaMA-3.1-8B-Instruct for 1 epoch at a 6e-5 learning rate. 
The specific ablation tests we included are 1) \textbf{4 hierarchies}: from a single document, we generated hierarchical ordering data using the algorithm specified in Section \ref{sec:coherent-single}. 2) \textbf{4 hierarchies with multi-hop reasoning}: In addition to the 4 hierachies set up in Section \ref{sec:coherent-single}, every time we generate a new QA pair, there is a 20 $\%$ chance that a multi-hop question-answer pair will follow. 3) \textbf{4 hierarchies without order}: hierarchical questions were generated without enforcing the order from Section \ref{sec:coherent-single}, testing if strict hierarchy enforcement improves outcomes. 4) \textbf{Diverse questions}: this setup generated various question types to test if diversity improves performance, as outlined in Section \ref{sec:coherent-single}. 

The results of these ablation studies on InfiniteBench are summarized in Table \ref{table:ablation-infinitebench}. The key findings include: 1) Multi-Hop Reasoning Improves Performance: Among all configurations, multi-hop reasoning achieved the highest average score of 54.70, demonstrating the importance of capturing cross-document relationships and broader reasoning capabilities. 2) Diverse Questions Provide Broad Improvements: The diverse questions setup achieved the second-highest score of 52.41, highlighting the value of introducing variety in QA generation for instruction-tuning data. 3) Hierarchical Ordering Boosts Performance: Both the strict hierarchical model (52.08) and the random hierarchical model (50.69) outperformed the base LLaMA-3.1-8B-Instruct (51.31), validating the effectiveness of hierarchical structuring, even when not strictly ordered. 

The LongBench results (presented in Table \ref{table:100Kablation-longbench}) provide additional insights, though the differences between configurations are relatively minor. This is likely because LongBench evaluates models on short contexts (up to 10K tokens), which do not fully leverage the strengths of hierarchical or multi-hop structures designed for longer contexts.
In summary, the ablation tests show that hierarchical ordering, multi-hop reasoning, and diverse questions are key to optimizing performance on long-context tasks. 

\begin{table}[t]
\centering
\small
\label{apx:longbench}
\caption{Ablation study on LongBench with 100K context length.}
\label{tab:model-performance-longbench-ablations}

\setlength{\tabcolsep}{4pt}
\begin{tabular}{@{}lcccccc@{}}
\toprule
\textbf{Task} & \textbf{\makecell{LLaMA-3.1-8B-\\Instruct}} & \textbf{\makecell{4\\hierarchies}} & \textbf{\makecell{4 hierarchies\\multi-hop}} & \textbf{\makecell{4 hierarchies\\random}} & \textbf{\makecell{diverse\\questions}} \\
\midrule
NarrativeQA & 25.48 & 25.89 & 25.10 & 25.04 & \textbf{27.91} \\
Qasper & 45.33 & \textbf{47.02} & 44.79 & 46.00 & 46.25 \\
MultiFieldQA-en & \textbf{54.98} & 54.86 & 53.96 & 54.86 & 53.75 \\
MultiFieldQA-zh & \textbf{61.83} & 55.75 & 54.87 & 59.89 & 56.14 \\
\midrule
Single Document & \textbf{46.91} & 45.88 & 44.68 & 46.45 & 46.01 \\
\midrule
HotpotQA & 55.00 & 56.67 & 56.91 & 55.83 & \textbf{58.34} \\
2WikiMQA & 44.95 & 52.19 & \textbf{52.96} & 48.74 & 52.71 \\
Musique & \textbf{31.76} & 29.15 & 28.55 & 29.85 & 28.10 \\
DuReader & 34.10 & \textbf{36.83} & 36.32 & 35.57 & 36.74 \\
\midrule
Multi-Document & 41.45 & 43.71 & 43.69 & 42.50 & \textbf{43.97} \\
\midrule
GovReport & 35.07 & 34.39 & 33.72 & 35.31 & \textbf{35.33} \\
QMSum & 25.13 & 25.15 & 25.27 & \textbf{25.52} & 25.38 \\
MultiNews & 27.08 & 27.34 & \textbf{27.48} & 27.29 & 27.46 \\
VCSUM & \textbf{17.10} & 16.12 & 16.75 & 16.13 & 16.40 \\
\midrule
Summarization & 26.10 & 25.75 & 25.81 & 26.06 & \textbf{26.14} \\
\midrule
TREC & 72.50 & \textbf{73.00} & \textbf{73.00} & \textbf{73.00} & 72.00 \\
TriviaQA & 91.65 & \textbf{92.28} & 92.25 & 91.87 & 91.83 \\
SAMSum & 43.77 & 43.81 & 43.98 & 44.49 & \textbf{45.48} \\
LSHT & 46.00 & 46.00 & 47.00 & 47.00 & \textbf{48.00} \\
\midrule
Few-shot Learning & 63.48 & 63.77 & 64.06 & 64.09 & \textbf{64.33} \\
\midrule
Passage Count & 6.55 & 4.00 & 3.00 & \textbf{7.56} & 5.00 \\
PassageRetrieval-e & \textbf{99.50} & 99.00 & 99.00 & 98.50 & 98.50 \\
PassageRetrieval-z & 96.38 & 98.50 & \textbf{100.00} & 94.63 & 99.50 \\
\midrule
Synthetic Tasks & 67.48 & 67.17 & 67.33 & 66.90 & \textbf{67.67} \\
\midrule
All & 48.11 & 48.31 & 48.15 & 48.27 & \textbf{48.67} \\
\bottomrule
\end{tabular}
\label{table:100Kablation-longbench}
\end{table}

\begin{table}[t]
\centering
\small
\caption{Ablation study on InfiniteBench with 100K context length.}
\label{tab:model-comparison}

\begin{tabular}{@{}lcccccc@{}}
\toprule
& \textbf{\makecell[c]{LLaMA-3.1-8B-\\Instruct}} & \textbf{\makecell[c]{4 hierarchies}} & \textbf{\makecell[c]{diverse\\questions}} & \textbf{\makecell[c]{4 hierarchies\\random}} & \textbf{\makecell[c]{4 hierarchies\\multi-hop}} \\
\midrule
Retrieve.PassKey & \textbf{100.00} & 86.66 & 86.66 & 86.66 & \textbf{100.00} \\
Retrieve.Number  & 95.33 & 86.66 & 86.00 & 85.33 & \textbf{96.66} \\
Retrieve.KV      & 42.66 & \textbf{60.00} & 58.00 & 58.66 &  57.33 \\
En.Sum           & \textbf{27.63} & 23.02 & 24.11 & 22.77 & 22.67 \\
En.QA            & 24.83 & 29.66 & \textbf{32.50} & 25.40 & 30.25 \\
En.MC            & 68.00 & 70.66 & \textbf{72.00} & 70.00 & 70.66 \\
En.Dia           & 16.66 & 24.66 & 23.33 & 20.66 & \textbf{26.00} \\
Math.Find        & 35.33 & 35.33 & \textbf{36.66} & 36.00 & 34.00 \\
\midrule
Average          & 51.31 & 52.08 & 52.41 & 50.69 & \textbf{54.70} \\
\bottomrule
\end{tabular}
\vspace{-1em}
\label{table:ablation-infinitebench}
\end{table}

\end{document}